\documentclass[10pt,journal,compsoc]{IEEEtran}

\ifCLASSOPTIONcompsoc
  \usepackage[nocompress]{cite}
\else
  \usepackage{cite}
\fi
\usepackage{booktabs}
\usepackage{graphicx}
\usepackage{float}
\usepackage{color}
\ifCLASSINFOpdf
\else
\fi
\usepackage{doi}
\usepackage{hyperref}

\usepackage{algorithm}  
\usepackage{algpseudocode}
\usepackage{amsmath}
\usepackage{multirow}

\usepackage{subfigure}
\begin{document}
%
\title{Emotional Contagion-Aware Deep Reinforcement Learning for Antagonistic Crowd Simulation}

\author{Pei Lv,
        Qingqing Yu,
        Boya Xu,
        Chaochao Li,
        Bing Zhou
        and Mingliang Xu 
\IEEEcompsocitemizethanks{\IEEEcompsocthanksitem Pei Lv, Qingqing Yu, Boya Xu, Chaochao Li, Bing Zhou and Mingliang Xu are with the School of Computer and Artificial Intelligence, Zhengzhou University, China, 450000.
\IEEEcompsocthanksitem The corresponding author is Mingliang Xu.
Contact e-mail: iexumingliang@zzu.edu.cn }
\thanks{Manuscript received April 19, 2005; revised August 26, 2015.}}

%
%


\markboth{}
{Shell \MakeLowercase{\textit{et al.}}: Bare Demo of IEEEtran.cls for Computer Society Journals}
%



\IEEEtitleabstractindextext{%
\begin{abstract}
The antagonistic behavior in the crowd usually exacerbates the seriousness of the situation in sudden riots, where the antagonistic emotional contagion and behavioral decision making play very important roles. However, the complex mechanism of antagonistic emotion influencing decision making, especially in the environment of sudden confrontation, has not yet been explored very clearly. In this paper, we propose an Emotional contagion-aware Deep reinforcement learning model for Antagonistic Crowd Simulation (ACSED). Firstly, we build a group emotional contagion module based on the improved Susceptible Infected Susceptible (SIS) infection disease model, and estimate the emotional state of the group at each time step during the simulation. Then, the tendency of crowd antagonistic action is estimated based on Deep Q Network (DQN), where the agent learns the action autonomously, and leverages the mean field theory to quickly calculate the influence of other surrounding individuals on the central one. Finally, the rationality of the predicted actions by DQN is further analyzed in combination with group emotion, and the final action of the agent is determined. The proposed method in this paper is verified through several experiments with different settings. The results prove that the antagonistic emotion has a vital impact on the group combat, and positive emotional states are more conducive to combat. Moreover, by comparing the simulation results with real scenes, the feasibility of our method is further confirmed, which can provide good reference to formulate battle plans and improve the win rate of righteous groups in a variety of situations.

\end{abstract}

}

\maketitle

\IEEEdisplaynontitleabstractindextext

%
\IEEEpeerreviewmaketitle

\IEEEraisesectionheading{\section{Introduction}\label{sec:introduction}}

%
%
%
%
\IEEEPARstart{W}{ith} the rapid development of the global economy and the growth of urban population, the frequency and severity of emergencies continue to rise. 
These emergencies have the characteristics of uncertainty, suddenness, and harmfulness. Once the event occurs, it may cause serious harm to society and citizens, and even irreversible consequences, e.g. the riots. For example, in August 2011, a demonstration in London suddenly turned into a violent confrontation\cite{2016Analysis}, where criminals attacked the police, innocent people, and destroyed public property. In September 2020, in Kentucky, the United States, due to conflicts between demonstrators from two different camps, the two sides gathered in public venue to confront and provoked each other, which aroused social concern. There are more and more similar sudden crowd incidents, and the harm and losses caused by each incident are incalculable and shocking. How to enable relevant departments to efficiently resolve such incidents has become a matter of great concern to all the society, and it is also one key issue that many scholars have devoted themselves to solve.


As an important research direction in the field of computer graphics, crowd simulation has been widely used in many fields such as security management, military exercises, and traffic planning. Especially in the face of riots, this kind of methods is often used to help simulate the corresponding process, which will save a lot of public resources compared to traditional solutions, such as manual exercises. By modeling and simulating the evolutionary process of crowd movement, it is possible to have more detailed understanding of the riots and their trend under emergencies, to truly reproduce such crowd behaviors. Then, we can quickly analyze and formulate effective decisions to quell the incident and reduce the potential loss. In the war game, researchers often deploy combat decisions in this way\cite{2018Bot}\cite{2018Simulation}\cite{2003Research}. However, in sudden real riots, there are many factors playing different roles when the crowd fights happen, and the emotions affect the antagonistic behavior in the group to a large extent~\cite{2009Evolutionary} in the way of decision-making of crowd behaviors\cite{2018Strategies}. Therefore, when planning antagonistic crowd behaviors, emotional factors must be incorporated into the crowd simulation model carefully.

In another aspect, individuals in antagonistic groups need to quickly make decisions in complex and changeable environment, and it is difficult for individuals to learn effective experience from similar events in the past. Although the existing crowd simulation models have considered the emotional factors when planning behaviors, it is not so reasonable to design the overall movement trend and specific behaviors of the agent in advance at the same time, and the formulated behaviors may deviate from the real cases. Recently, deep reinforcement learning that combines the perception capabilities of deep learning and decision-making capabilities of reinforcement learning have been widely explored. Many researchers try to use this novel technique to study the decision-making behavior of the crowd to plan the decision-making better ~\cite{2019Agent}\cite{16}. However, when using this method to model the crowd behaviors, the attributes of agent, such as the inherent group emotions, are not fully considered. The constructed simulation model lacks authenticity, and the win rate of the righteous side is not high enough during the battle.

To solve the above problems, this paper proposes an antagonistic crowd behavior simulation model (ACSED) integrating emotional contagion into deep reinforcement learning. First of all, we fully consider the key role of emotions on antagonistic crowd behaviors. The antagonistic emotional contagion module is built based on the improved SIS model, and combines with specific combat situations to estimate individual emotions (Section 3.1). Then, we use deep reinforcement learning to construct antagonistic action predict module for different groups, allowing the agent to learn decision-making action efficiently and autonomously, and leveraging the mean field theory\cite{stanley1971phase} to simplify the calculation complexity  (Section 3.2). Finally, we combine with the emotion of each agent in the crowd to judge whether the learned action is reasonable, and the final battle is determined according to the action rules of agent under different emotional states (Section 3.3).

The simulation experiments prove that the proposed method is helpful for studying the antagonistic behaviors in the riots, and able to formulate more realistic and reasonable combat plan for the righteous group and improve the win rate. The main contributions of this paper is as following:
\begin{itemize}
\item We propose an emotional contagion-aware deep reinforcement learning model for antagonistic crowd simulation (ACSED). The DQN and mean field theory are introduced to predict the action in the crowd. The proposed model can provide the agents with more reasonable and effective actions.

\item We introduce an antagonistic emotional contagion module to calculate individual emotions and formulate the behavioral rules of agents under different emotions. This module fully considers the influence of emotions on the intensity of the attack, and establishes a connection between the individual's emotions and the suffered harm to improve the authenticity of the simulation.

\item We develop an antagonistic action prediction module to estimate the potential action for each agent in the crowd. Meanwhile, the emotion of the agent is used to further analyze whether the predicted action is reasonable, and combined to determine the final action better. 

\item Our model is able to improve the win rate of the battle in crowd antagonistic scene efficiently, and fully explore the advantages of the emotion to win more with less, which is more in line with the realistic cases.
\end{itemize}

The rest of this paper is organized as follows. The second section discusses the related work involved in this paper. The third section proposes the antagonistic crowd simulation model. The rationality of the proposed model is verified through different experiments in the fourth section. Finally, the conclusion including a summary and future work is demonstrated in the last section.

\section{Related Work}

\subsection{Crowd simulation}
Crowd simulation \cite{2017Crowd} is of great significance to many fields and is widely used in the scene such as public safety, military training, and video games\cite{2014Crowd}. It can be generally divided into macro models and micro models~\cite{4}. Macro models regard the crowd as a whole, focusing on the movement trends of the whole crowd, and the details of individual movement is relatively rough. Representative methods include aggregate dynamics models~\cite{2009Aggregate}, potential field models~\cite{patil2010directing}, and so on. Micro models pay attention to the concrete details of individual movement in the crowd, studying their behavioral rules and decision-making process. This kind of model can more realistically show the complex interactions among individuals, which is also used in this paper.

Classical microscopic models include cellular automata model \cite{5}, social force model \cite{6}, multi-agent model \cite{7} and so on. Ren et al. \cite{8865441}proposed a combined multi-agent model and data-driven approach for heterogeneous group simulation, where they estimate crowd motion states from a real dataset including position, velocity, and control direction information. Sahil et al.\cite{8642370} inferred user intent based on the observed proxemics and gaze-based cues, and the inferred intent is used to guide the response of the virtual agent and generate locomotion and gaze-based behaviors in shared avatar-agent virtual environments. Recent years, plenty of researchers have devoted themselves to the simulation of crowd emergency or riots with emotion factor. Beltaief et al. \cite{13} proposed a multi-agent simulation model based on psychological theory, which was able to simulate the crowd gathering phenomenon more realistically. Spartalis et al.\cite{8} used cellular automata to simulate pedestrian movement, and introduced group categorization and guidance attributes to help study the influence of guidance on the crowd evacuation. The crowd is categorized according to emotion and a special group has leadership characteristics. Pax et al.\cite{11} built an agent-based architecture, which allows for the efficient simulation of indoor scene without losing the ability to specify rich and heterogeneous agent behaviours. 
With the rapid development of multi-agent deep reinforcement learning~\cite{14}, researchers begin to use this novel tool to model the complex action of agents. Especially for emergencies. This kind of method can estimate actions in a closer way to human thinking, which is more reasonable and reliable than previous methods. Gupta et al. \cite{15} used the multi-agent deep reinforcement learning to explore the cooperative strategy learning problem by the complex and partly observable agents. By designing a set of experiments for cooperative control tasks, the effectiveness of their method was proved. Zhang et al.\cite{9437632} proposed a data-driven crowd evacuation framework based on hierarchical deep reinforcement learning. In the micro-control layer, the track sequence learned in the macro-control layer is used as the motion target, and the multi-agent deep reinforcement learning method is used to learn the collision-free motion velocity of the individual.

Although the crowd simulation have made great progress up to now, they still face enormous challenges to simulate the antagonistic crowd behaviors. One of the most important reasons is that the antagonistic behaviors are more complex and easily changeable caused by unstable emotion, which bring great challenges to plan accurate agent actions in advance based on previous experience. In another aspect, although the usage of multi-agent deep reinforcement learning allows agents to learn action efficiently, this novel technology has not touched the inevitable emotion factor in crowd simulation, and the authenticity of simulation still needs to be improved.

\subsection{Emotional contagion}
Emotion is a short-term psychological state produced by individuals based on subjective cognition, which is closely related to feelings, thoughts, and actions \cite{1988The}. When individuals receive external stimuli, their emotions will change, and then the emotions cause changes in human behavior. Individual emotion plays a vital role in the process of their behavioral decision-making, and will have a great influence on their behaviors \cite{durupinar2015psychological}. The emotional cognitive theory believes that emotions arise from an individual’s evaluation of something that is beneficial or harmful to oneself \cite{2020Cognitive}. 
Emotional contagion is one very typical and essential factor in the crowd movement. Therefore, during the crowd simulation, we should not only pay attention to the emotional state of the individual, but also need to study the process of emotional contagion\cite{2016Dynamic}. Emotional contagion methods are mainly divided into two categories. One is based on thermodynamic methods \cite{19} and the other is based on epidemiological methods \cite{20}. 
The epidemiological methods are inspired by the spreading mechanism of epidemic diseases, where susceptible persons are at risk of being infected when they come into contact with infected one. 

Hill et al. \cite{21} proved that the epidemic model can be used to study the problems related to emotional contagion in the population. They introduced a new form of the classic SIS (Susceptible Infected Susceptible) model that includes the possibility of ``spontaneous" (or ``automatic") infection, termed the SISa model. On the basis of this work, Liu et al. \cite{22} proposed SOSa-SPSa model, which divided crowd emotions into positive and negative states, and further studied the internal mechanism of emotional contagion. Nizamani et al. \cite{Galam2014From} constructed an emotional contagion model based on the spreading epidemics models, divided the population into five categories, and studied the spread of anger among groups. Zhao et al. \cite{24} constructed an improved SIRS model to study the spread of panic in subway passengers, and analyzed in detail the influence of crowd density and individual psychology on group emotions.  Mao et al.\cite{9} proposed one crowd simulation approach involving the OCEAN personality model and the OCC emotion model, and combined the CA-SIRS emotional contagion model, to simulate diverse crowd behaviors. Li et al.\cite{li2019acsee} proposed a crowd antagonistic behavior simulation model (ACSEE) by combining adversarial emotions and evolutionary game theory. They used cellular automata to determine the position of the agent, and simulated similar antagonistic crowd behavior in real scenes. Mao et al.\cite{mao2020personality} proposed a unified framework to simulate emergency evacuation in virtual environments. The emotional contagion in their work is considered from three aspects: intra-group contagion, inter-group contagion and emotional contagion based on the third-party authority. Xu et al.\cite{8667341} proposed a novel approach for crowd evacuation simulation by modeling panic generation and contagion in multi-hazard situations.

Inspired by above work, aiming at the spread of emotions between groups in the outbreak of sudden riot, we build an antagonistic emotional contagion module based on the improved SIS model, to explore the influence of emotions on antagonistic crowd behavior, and formulate more realistic actions of each agent in the crowd.

\subsection{Deep reinforcement learning}

In recent years, deep reinforcement learning (DRL) has become an important research direction in the field of artificial intelligence\cite{2016Deep}. It is widely used in many important areas such as behavioral decision making, robot control, parameter optimization, etc. DRL integrated deep learning with Reinforcement Learning (RL) to partially overcome the curse of dimensionality. DeepMind proposed the DQN algorithm \cite{2013Playing}, which combined Q-learning with deep learning together. The state and action were regarded as the input of deep neural network. Wang et al.\cite{wang2019improved} proposed an improved multi-agent reinforcement learning method, combining with an improved social force model in crowd evacuation simulation. Toghiani-Rizi et al. \cite{toghiani2017evaluating} evaluated the ability of three deep reinforcement learning algorithms to learn the tasks in simulated ground combat scene, proving that deep reinforcement learning has the potential to improve practices and techniques for modeling tactical behavior. Yang et al.\cite{16} combined Q-learning and mean field theory to propose the Mean Field Q-learning (MF-Q) algorithm, which was dedicated to solving the problem of large-scale agent learning with higher calculation efficiency. In detail, when they calculated the influence on certain agent, a mean value was introduced to replace the effect of all other agents, which greatly simplified the increasing model space due to the increase of the number of agents. The effectiveness of the MF-Q algorithm is verified in a simple crowd antagonistic scene and can guarantee the win rate of group battles.


However, original DRL-based algorithms do not fully consider complex factors that affect individual actions in different groups, such as emotions. In this paper, we try to improve the work of \cite{16}, through integrating emotions with multi-agent deep reinforcement learning to build a new simulation model of antagonistic crowd behavior.

\begin{figure*}[htp]
\centering
\includegraphics[height=8.8cm,width=17.7cm]{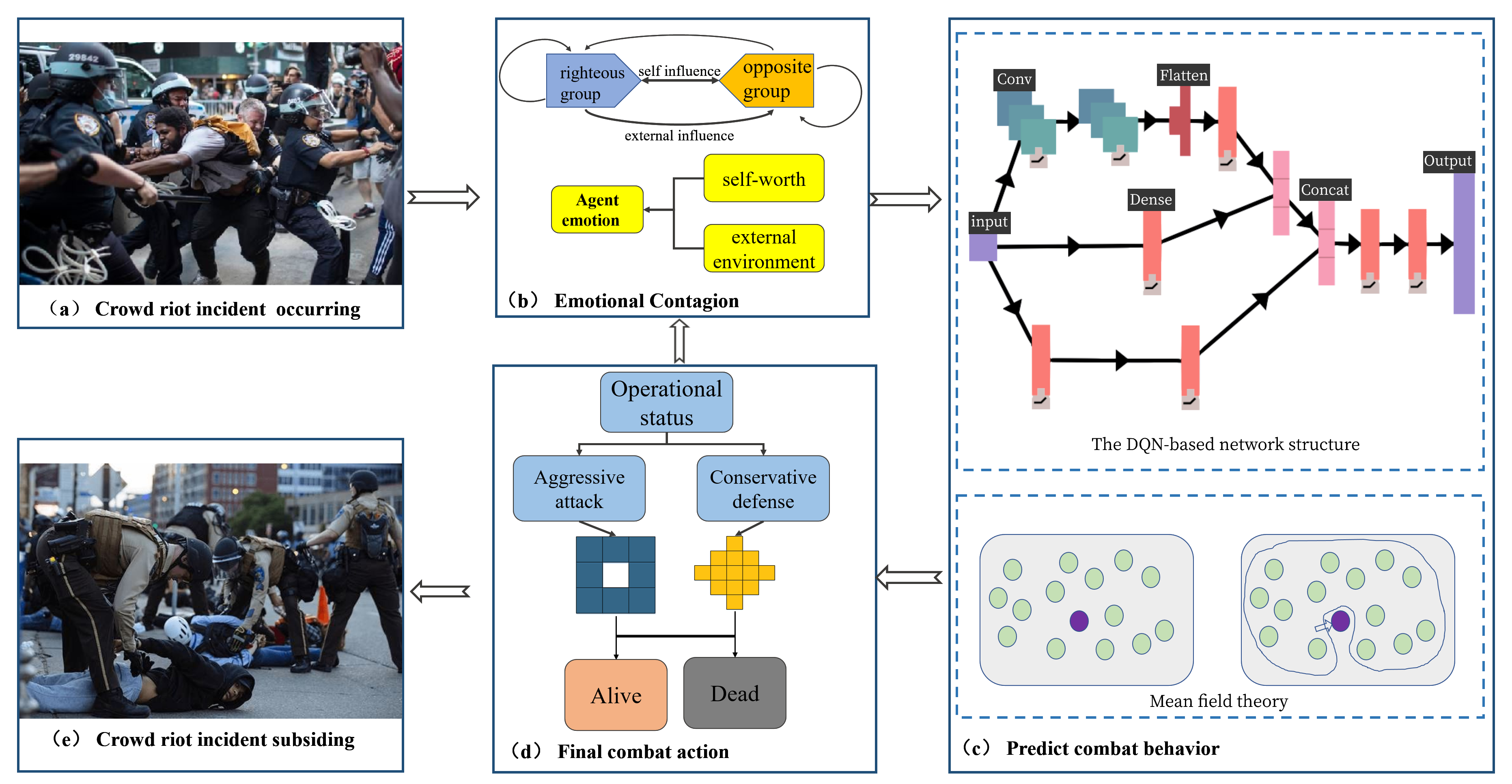}
\caption{Overview of our ACSED model: (a) The group who provokes the riot event is called the opposite side, and the group who puts down the riot event is called the righteous side. Individuals in the riot can perceive the environmental information. (b) We calculate the emotional state of individuals by the emotional contagion module consisting of two parts: external influence and self-influence. The emotion of each agent will be updated at each time step. (c) One neural network model is built based on DQN and mean field theory to predict the initial action of individual, which includes the attack and the move. (d) The final action of one agent is determined according to the action rules under different emotional states. The reward of each time step acts as a self-influence to affect the agent's emotion. (e) If the crowd violence subsides, the incident ends; otherwise, the process will be repeated from the step (b).}
\label{pic1}
\end{figure*}

\section{Antagonistic Crowd Simulation Model}
This paper mainly studies the riots and proposes an emotional contagion-aware deep reinforcement learning model for antagonistic crowd simulation. We define the side that provokes the riot as the opposite, and the side that calms the situation as the righteous. By studying the characteristics and laws of crowd movement in riot scenes, we can formulate reasonable and efficient decision-making actions for the righteous group and improve their win rate.

The framework of the antagonistic crowd behavior simulation model is shown in Figure \ref{pic1}. Firstly, based on the improved SIS model, we combine with the combat situation to build an antagonistic emotional contagion module to calculate the emotions of different groups in emergencies. Then, we propose an antagonistic action prediction module based on DQN and mean field theory, and use these novel tools to reasonably analyze and predict the agent action in the crowd. Finally, the final action of the agent is determined according to the action rules under different emotional state. The method proposed in this paper helps to study the antagonistic behavior of the crowd under violent and terrorist incidents, so as to formulate more conducive action plans for the righteous group.

\subsection{Antagonistic Emotional Contagion Module}
The current methods based on deep reinforcement learning to model antagonistic crowd behavior do not fully consider the emotional factors of the individual and there are some disadvantages such as deviations between the simulation results and the ground truth, and the unsatisfactory action. To solve the above problems, we build an emotional contagion module which is more suitable for confrontation scene. This module is based on the improved SIS model and combines with the combat situation to analyze the specific effect of emotions on antagonistic crowd behavior. 
The decision can be more realistic, reasonable and credible.

We divide the emotions of agents into positive emotions and negative ones. According to the warehouse model in the epidemic model, the population is divided into susceptible and infected. $E_{i}$ represents the emotion intensity of $Agent_{i}$, which is set in the range $[0,1]$ for the righteous and $[-1,0]$ for the opposite. The larger the emotional value of both groups, the more positive they are, on the contrary, the smaller the more negative. When it is close to the median value of $0.5$ or $-0.5$, it means that $Agent_{i}$ is in a peaceful state. For different types of agents, the more positive the emotional state of the righteous side, the more daring to take offensive action to subdue the opposite side. On the contrary, they will fear the opposite side and fight passively. The more negative the emotional state of the opposite side, the more inclined to challenge the righteous side and will attack them actively.

Emotional contagion between groups is similar to the spreading process of infectious diseases. Individual emotions will not only be affected by other people in the environment, but also by their own. Therefore, we calculate the agent's emotion from the external environment and self evaluation. The first part is the external influence $E_{i}^{e x}$. The influence of the external environment comes from two sources, including the distance between $Agent_{i}$ and surrounding agents and the emotions of surrounding agents. The second part is the self-influence $E_{i}^{se}$. The influence of self-assessment refers to the influence of the behavioral value assessment obtained by $Agent_{i}$ on its emotions. According to the emotion cognitive evaluation theory, emotions are generated from the evaluation of some specific aspects between the individual and environment, thereby generating an adaptive response to the current situation. The calculation of emotional contagion is shown in Formula \eqref{equ1}:


%

\begin{equation}
\label{equ1}
E_{i}=E_{i}^{e x}+E_{i}^{se}
\end{equation}

First, we calculate the amount of changes in the emotion of $Agent_{i}$ after being affected by the external environment, that is, when $Agent_{i}$ interacts with other agents around it, it will be affected by the emotions of other agents. Inspired by \cite{25}, the changing values of emotional contagion of $Agent_{i}$ is defined in Formula \eqref{equ2}:

\begin{equation}
\label{equ2}
\Delta E_{i, j}^{e x}(t)=[1-\frac{1}{1+exp(-D)}] \times E_{i}(t) \times A_{j, i} \times B_{i, j}
\end{equation}
where $D$ represents the distance between $Agent_{i}$ and other $Agent_{j}$, $E_{i}$ represents the emotion of $Agent_{i}$, $A_{j,i}$ is the intensity of emotion received by the affected $Agent_{i}$ from the influencing $Agent_{j}$, and $B_{j, i}$ refers to the emotional intensity sent from $Agent_{j}$ to $Agent_{i}$. The external emotional contagion is the result of the contagion of the righteous and opposite agents in the perceiving range on $Agent_{i}$. People who belong to the same team as  $Agent_{i}$ will have a positive effect on their emotions, otherwise they will have a negative effect. Formula \eqref{equ3} is to calculate the external emotional contagion of the righteous at time $t$. Formula \eqref{equ4} is to calculate the external emotional contagion of the opposite at time $t$.
\begin{equation}
\label{equ3}
\Delta E_{r}^{e x}=\sum_{i=1}^{m} \Delta E_{r, r_{i}}^{e x}(t)+\sum_{j=1}^{n} \Delta E_{r, o_{j}}^{e x}(t)
\end{equation}
\begin{equation}
\label{equ4}
\Delta E_{o}^{e x}=\sum_{i=1}^{n} \Delta E_{o, o_{i}}^{e x}(t)+\sum_{j=1}^{m} \Delta E_{o, r_{j}}^{e x}(t)
\end{equation}

When we calculate the emotional state of one agent, it is necessary to consider the influence of the agent on the emotions of itself and others after taking actions. Inspired by \cite{li2019acsee}\cite{10.1007/BF01238028}, we calculate the influence of self-evaluation on the emotion of the agent based on the reward value in reinforcement learning. During the battle between two groups, the agent will obtain the corresponding reward value after taking the action, which is used to evaluate the performance of the agent. The mental emotion calculation method is as follows:
\begin{equation}
\label{equ5}
\Delta E_{i}^{se}(t)=0.1 \times\left(\frac{1}{\delta+\exp \left(\gamma / r_{i}(t)\right)}\right), r_{i}(t) \geq \gamma
\end{equation}
\begin{equation}
\label{equ6}
\Delta E_{i}^{se}(t)=-0.1 \times\left(\frac{1}{\delta+\exp \left( r_{i}(t) /\gamma\right)}\right), r_{i}(t) \leq-\gamma
\end{equation}
where $r_{i}(t)$ represents the difference between the reward values of two consequent time steps, $\delta$ is an empirical parameter. When $r_{i}(t) \in(-\gamma, \gamma)$, the action of $Agent_{i}$ has less effect on its emotions and can be ignored. When $r_{i}(t) \geq \gamma$, it means that $Agent_{i}$ performs the action to promote the battle result. If $Agent_{i}$ is righteous, its emotions will become positive, otherwise if it is opposite, it will become negative. When $r_{i}(t) \leq-\gamma$, it means that the action performed by $Agent_{i}$ is not conducive to the current combat situation. If $Agent_{i}$ is positive, its emotions will become negative, and if it is negative, it will become positive. We calculate the emotional contagion of $Agent_{i}$ according to Formula \eqref{equ5}\eqref{equ6}. According to above formulas,we calculate the amount of emotional contagion of $Agent_{i}$.

\begin{equation}
\label{equ7}
E(i, t)=E(i, t-1)+\Delta E_{i}^{ex}(t)+\Delta E_{i}^{se}(t)
\end{equation}

At time $t$ , the emotional value is calculated according to Formula \eqref{equ7}, and the emotional value of $Agent_{i}$ at time $t-1$ is summed with the increase in emotional contagion obtained by $Agent_{i}$ at time $t$.

\subsection{Antagonistic Action Prediction Module}


In this section, we build an action prediction module based on DQN and mean field theory, which is able to calculate the antagonistic action taken by one agent efficiently. There are two important reasons for choosing DQN. On the one hand, riots are usually sudden and have different types, and it is difficult to accumulate experience from past events. DQN is suitable for solving such problems with less prior knowledge. On the other hand, DQN belongs to the off-policy model with experience playback pool, which can break the correlation between existing data and realize a more stable learning process. At the same time, the usage of convolutional neural network as a value function can fit the Q table in the Q-learning algorithm better. 

The overall pipeline of this module is as following. Firstly, the two sides use the action provided by the initial network to fight against each other, and the sampled data will be stored in the experience replay pool. Then, we randomly sample data from the experience replay pool to train the network and iterate for many rounds. Finally, the trained model is used to predict the initial action of each agent.

\begin{figure}[htpb]
\centering
\includegraphics[height=5.5cm,width=8.0cm]{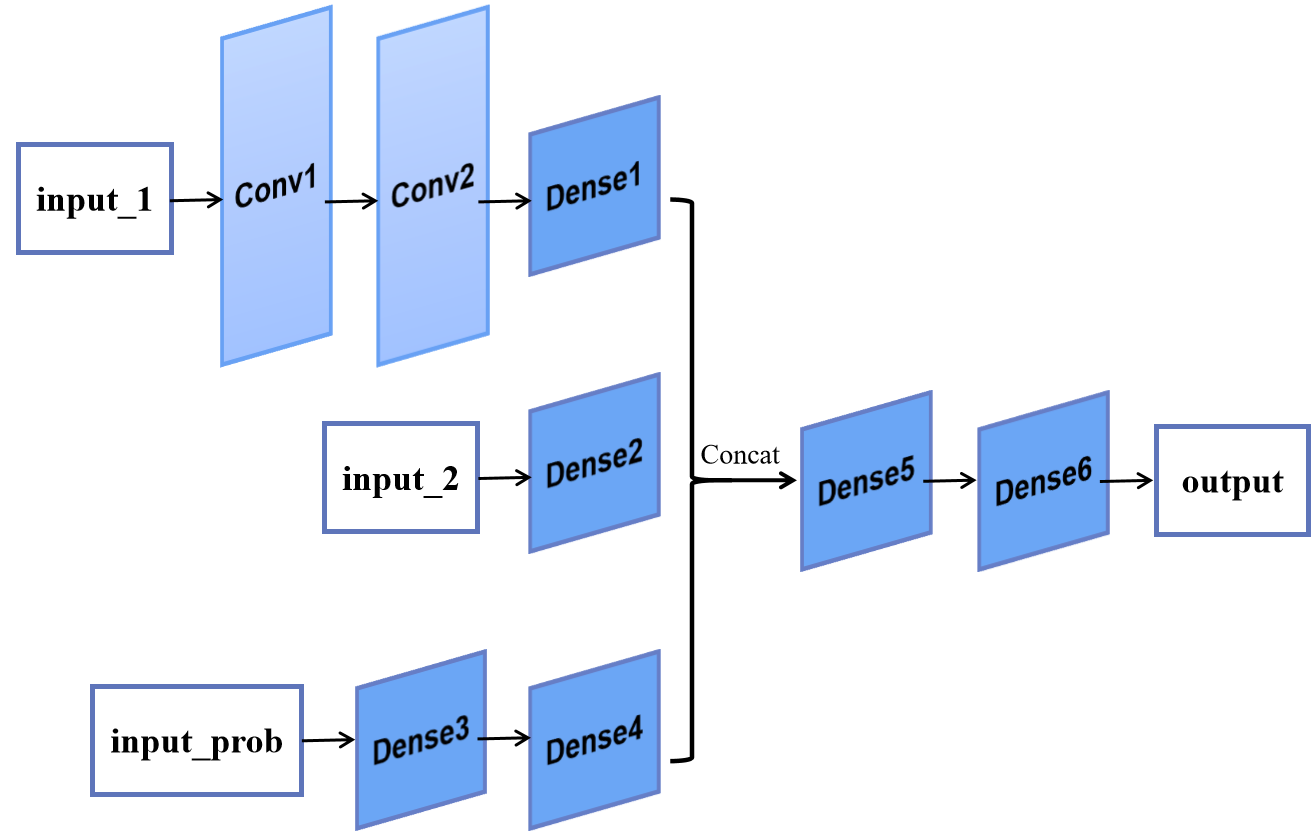}
\caption{The DQN-based network structure in ACSED which consists of two convolutional layers and six fully connected layers, and the network outputs the Q value.}
\label{pic3}
\end{figure}

The specific network structure is shown in Figure \ref{pic3}. input\_1 is used as the input of the first convolution layer and contains information about the categories of other agents within the perceiving range of certain agent. Then, the output of two convolutional layers is input to the first fully connected layer. input\_2 is used as the input of the second fully connected layer, which contains information such as ID information, location, action, and emotional value. input\_prob is the mean action, and we input it into the fully connected layer. The Q value is obtained through the output layer, and the decision-making action of the agent at the next moment is determined according to the Q value.

The mean action mentioned above is calculated using mean field theory. Due to the large number of agents involved in riots, the calculation complexity must be simplified while constructing the corresponding network \cite{16}. Inspired by \cite{8}, we approximate all the influence of the neighboring agents as one influence, and the actions of the neighboring agents as an action. 

The dimension of joint action $a$ grows proportionally with the number of agents $M$. Since all agents act strategically and simultaneously to evaluate their value functions based on joint actions, the learning of standard Q-function $Q^{i}(s, a)$ becomes infeasible. To solve this problem, we factorize the Q-function using only pairwise local interactions. The action of one agent and that of its neighbors can be combined as an action pair in Formula \eqref{equ8}.
\begin{equation}
\label{equ8}
Q^{i}(s, a)=\frac{1}{M^{i}} \sum_{k \in M\left(i\right)} Q^{i}\left(s, a^{i}, a^{k}\right)
\end{equation}
$M\left(i\right)$ represents the index set of the neighboring agents of agent $i$ with the size $M^{i}=\left|M\left(i\right)\right|$. $ a^{i}$ is the action taken by $Agent_{i}$ in the state $a$, and $ a^{k}$ represents the actions of other neighboring agents. The pairwise approximation of agents and their neighbors not only reduces the complexity of interactions among agents, but still implicitly preserves the global interactions between any pair of agents \cite{1993The}.

The $Q^{i}\left(s, a^{i}, a^{k}\right)$ in Formula \eqref{equ8} can be approximated using the mean field theory. When calculating the agent action at time $t$, the action estimated by DRL network at time $t-1$  will also be considered. The action of $Agent_{i}$ is a discrete action categorical variable represented by one-hot encoding. Through the action coding of the neighborhood agents, their mean action can be obtained in Formula \eqref{equ9}.

\begin{equation}
\label{equ9}
a^{k}=a^{-i}+\delta a^{i, k}, \text { where } a^{-i}=\frac{1}{M^{i}} \sum_{k} a^{k}
\end{equation}
$ a^{-i}$ represents the mean action. $\delta a^{i, k}$ is a small fluctuation value. In this formula, the Q function of  $Agent_{i}$ can be shown in Formula \eqref{equ10}.
\begin{equation}
\label{equ10}
Q^{i}(s, a)=\frac{1}{M^{i}} \sum_{k \in m} Q^{i}\left(s, a^{i}, a^{k}\right)=Q^{i}\left(s, a^{i}, a^{-i}\right)
\end{equation}

The mean action is taken as the key factor affecting the action of one agent, and this factor is taken as the input of the neural network to estimate the next action taken by the agent.

The way to update the Q value is as follows:

\begin{equation}
\label{equ11}
\begin{split}
Q_{t}^{i}\left(s, a^{i}, a^{-i}\right) =(&1-\alpha) Q_{t-1}^{i}\left(s, a^{i}, a^{-i}\right) \\
&+\alpha\left[r^{i}+\gamma v_{t-1}^{i}\left(s^{\prime}\right)\right]
\end{split}
\end{equation}

\begin{equation}
\label{equ12}
\begin{split}
v_{t-1}^{i}\left(s^{\prime}\right)=&
\sum_{a} \pi_{t-1}^{i}\left(a^{i} \mid s^{\prime}, a^{-i}\right) \\&E_{a^{-i}\left(a^{-i}\right) \sim \pi_{t-1}^{-i}}\left[Q_{t}^{i}\left(s, a^{i}, a^{-i}\right)\right]
\end{split}
\end{equation}
$Q_{t}^{i}\left(s, a^{i}, a^{-i}\right)$ equals the actual Q value obtained at time $t-1$ plus the maximum possible reward obtained at time $t$. The maximum possible reward includes the reward obtained by executing the current action and the maximum Q value that may be obtained at time $t$. $\alpha$ represents the learning rate. $\gamma \in(0,1)$ represents the discount factor, which is used to balance the relationship between short-term and future rewards. $v_{t-1}^{i}\left(s^{\prime}\right)$ is the mean field value function, and $\pi$ represents a random strategy.

The following loss function is used to train the network.
\begin{equation}
\label{equ13}
y_{i}=r^{i}\left(s, a^{i}, a^{-i}\right)+\gamma v_{t}^{i}\left(s^{\prime}\right)
\end{equation}

\begin{equation}
\label{equ14}
L\left(\phi^{i}\right)=\left(y_{i}-Q_{\phi^{i}}\left(s, a^{i}, a^{-i}\right)\right)^{2}
\end{equation}

The adaptive moment estimation method is used to reduce the error between the estimation of the Q network and the expected target value. Then, we use the method to minimize the loss function and update the network parameters in reverse, to continuously improve and optimize the original network. The trained model is saved and applied to the actual situation to predict the agent's action. We adopt the reward setting: -0.005 for every move, 0.2 for attacking an enemy, 5 for subduing an enemy, -0.1 for attacking an empty grid, and -0.1 for being attacked.

\subsection{Antagonistic Action Determination Module} 
In riots, the action of one individual is easily affected by its own emotion and others', and individuals with different emotional states tend to adopt different actions. This section will introduce how to leverage the initial action predicted by DRL with current emotional states to determine the action of the agent and improve the final win rate of the righteous group. The overall procedure of our algorithm is presented in Algorithm \ref{alg::antagonistic}.

\begin{algorithm}[htp]
  \caption{ACSED algorithm}
  \label{alg::antagonistic}
  \begin{algorithmic}[1]
     \State Initialize the attributes of $Agent_{i}$, such as ID information $id$, location $pos$, action $act$, emotional value $emo$, mean action $act_{prob}$, environmental awareness $view$;
     \State Initialize max time step $t=400$, discount rate $\gamma=0.95$, learning rate $lr=1e-4$, batch size $bs=256$, memory size $ms=2^{10}$;
    \While{$t<400$}
      \If{the size of any one team $live_{ct}<2$}
              \State break
      \Else
               \State Enter the parameters into the network to estimate 
                       \Statex \quad \qquad  the $Q$ value;
                \State Predict the action $acts_{p}$ based on the
                        maximum \Statex \quad \qquad $Q$ value;
                \State Update $pos$, $view$, and calculate the rewards $r$;        
                \State Obtain the agent’s final decision-making action           \Statex \quad \qquad $acts_{f}$ through specific action rules;
                \State Calculate the mean action $act_{prob}$ at next time        \Statex \quad \qquad step $t'$ according to the action $acts_{f}$ of the agent;
                \State Update the emotional value $emo'$ of the agent \Statex \quad \qquad at next time step $t'$ based on following function: \Statex \quad \qquad $E(i, t')=E(i, t)+\Delta E_{i, j}^{e s}(t')+\Delta E_{i}^{e r}(t')$;
      \EndIf
    \EndWhile
  \end{algorithmic}
\end{algorithm}

According to the behavioral tendency of agents and the type of actions under different emotions, a threshold $T$ is defined, and the combat state of agent is divided into two categories: aggressive offense and conservative defense. The behavioral tendency of agents in different situations is shown in Table \ref{table2}. If the agent is in an aggressive state, both the righteous and the opposite sides will be more proactive in taking offensive actions and attacking the opponent. At this time, the righteous agent have positive emotion, while the opposite will be negative. If the agent is in a conservative state, it means that the current state is not active. The agent only has a defensive mindset and prefers to adopting move-type action. At this time, the righteous have negative emotion, while the opposite will be positive. The positivity of the emotion of the opposite is reflected in the weak combat consciousness.

\renewcommand{\arraystretch}{1.2}
\begin{table}[!t]
\centering  
\caption{The behavioral tendency in different situations.}  
\label{table2}

\begin{tabular}{cccc}
\toprule[1pt]
Emotion & \begin{tabular}[c]{@{}c@{}}Combat\\ state\end{tabular}   & \begin{tabular}[c]{@{}c@{}}Agent \\action   \end{tabular}   & \begin{tabular}[c]{@{}c@{}}Emotional state\end{tabular}       \\ \midrule
$1>\left|E_{i}\right|>T>0$   & \begin{tabular}[c]{@{}c@{}}Aggressive\\ offense\end{tabular}   & \begin{tabular}[c]{@{}l@{}}attack
\end{tabular}   & \begin{tabular}[c]{@{}c@{}}The righteous\\ are positive;\\ the opposite\\ are negative\end{tabular}\\
\midrule$0<\left|E_{i}\right|<T<1$    & \begin{tabular}[c]{@{}c@{}}Conservative\\ defense\end{tabular} & \begin{tabular}[c]{@{}l@{}}move\end{tabular} & \begin{tabular}[c]{@{}c@{}}The righteous\\ are negative;\\ the opposite \\are positive\end{tabular}\\ 

\bottomrule[1pt]
\end{tabular}

\end{table}

We analyze whether the action predicted by the DRL network is reasonable according to the emotional state of the agent, so as to determine its final action. If the predicted action of the agent is unreasonable, it will adversely affect its own battle situation. Therefore, it is necessary to re-plan more advantageous actions according to the agent's combat state. As shown in Table\ref{table2}, the actions of the agent are divided into two categories, including the move action and attacking one. The detailed discussion will be divided into the following situations.

Firstly, if the action predicted by the network for $Agent_{i}$ belongs to the type of attack, the rationality of the action needs to be analyzed based on its current combat state and attack target.

If $Agent_{i}$ is currently aggressive:
\begin{itemize}
    \item \textbf{The target of $Agent_{i}$ is a blank location or wall.} Such an attack target is meaningless, so $Agent_{i}$ will choose another conservative agent to attack from nearby reachable targets. If there are multiple eligible agents, the closest agent will be selected.
    \item \textbf{The target of $Agent_{i}$ is the opposite $Agent_{j}$.} Firstly, the emotional value of $Agent_{i}$ and that of the attack target will be compared. If $\left|E_{i}\right|>\left|E_{j}\right|$ or $\left|\left|E_{i}\right|-\left|E_{j}\right|\right|<E_{-} t h$ ($E_{-} t h$ represents the emotional threshold), $Agent_{i}$ will execute the predicted action; if $\left|E_{i}\right|<\left|E_{j}\right|$ and $\left|\left|E_{i}\right|-\left|E_{j}\right|\right|>E_{-} t h$, it is necessary to determine the number of partners and that of opponents within the perceiving range of $Agent_{i}$. If the number of partners is less than that of opponents, the $Agent_{i}$ will choose the action corresponding to the largest Q value in the move-type. If the number of partners is more than the opponent, the agent with the smallest absolute emotional value in the opponent group will be attacked.
\end{itemize}

If $Agent_{i}$ is currently conservative:

\begin{itemize}
    \item {\bf The attack target of $Agent_{i}$ is a blank location or a wall.} $Agent_{i}$ will choose an action with the largest Q value among the move-type actions.
    \item {\bf The target of $Agent_{i}$ is the opposite agent.} We compare the emotional value of $Agent_{i}$ and that of the attack target.  If $\left|E_{i}\right|>\left|E_{j}\right|$ or $\left|\left|E_{i}\right|-\left|E_{j}\right|\right|<E_{-} t h$, we execute the attack action. If $\left|E_{i}\right|>1/2 T$, the attack action corresponding to the second largest Q value in the attack-type will be taken, and the attack target will be changed. If $\left|E_{i}\right|<1/2 T$, $Agent_{i}$ will choose an action with the largest Q value among the move-type actions.
\end{itemize}

Secondly, if the network predicts that the next action of $Agent_{i}$ belongs to the type of move, it is necessary to discuss the rationality of the action based on the current state of $Agent_{i}$ and the surrounding environment.

\begin{itemize}
    \item  {\bf If $Agent_{i}$ is currently aggressive.} We compare the emotional value of $Agent_{i}$ with that of all opponents within its perceiving field. If all opponent members within the perceiving range are more aggressive than $Agent_{i}$, then they will execute the move action. If there is an agent who is conservative in the opponent group, $Agent_{i}$ will attack it. When there are multiple eligible agents, the agent with the most conservative combat state will be selected to attack.
    \item {\bf If $Agent_{i}$ is currently  conservative.} It performs the predicted action.
\end{itemize}

In addition, we also consider the effect of opponent's emotion on the damage suffered by $Agent_{i}$. In fact, the attack intensity highly depends on the corresponding emotion when the agent takes offensive action. The more aggressive the combat state, the stronger the attack, and the damage received by $Agent_{i}$ after being attacked increases with the strength of the attack. 
Therefore, we associates the emotion of the agent with its health value based on above facts. When modeling crowd behaviors, the attributes of the agent are set based on the actual situation to ensure that the calculation results are in line with reality and increase the practicability of the model. Specifically, according to the emotional value when the agent takes the attack action, the damage of the attacked target is calculated, as shown in Formula \eqref{equ15}:

\begin{equation}
\label{equ15}
H p_{j}(t)=H p_{j}(t-1)-\beta \log _{\frac{1}{2}}\left(1-\left|E_{i}(t-1)\right|\right)
\end{equation}
$E_{i}(t-1)$ represents the emotion of  $Agent_{i}$. $\beta$ is an empirical coefficient, which is specifically set according to the experimental results. If $Agent_{i}$ is righteous, the more positive the emotion, the greater the intensity of the attack, and the more harm the attacked individual will suffer. If $Agent_{i}$ is the opposite, the more negative the emotion, the greater the intensity of the attack, and the more harm the attacked individual will suffer. $H p_{j}(t-1)$ represents the health value of $Agent_{i}$ at time $t-1$. When $H p_{j} \leq 0$ , the $Agent_{i}$ is subdued and does not have combat capability.

\section{Experimental Results}
Our experiment is implemented on Intel CPU i7-8700K, 3.70 GHz, 32GB memory, Linux operating system environment. C++ and Python language is used to realize the antagonistic crowd behavior simulation model. We verify the proposed method on MAgent, a multi-agent reinforcement learning platform that supports hundreds to millions of agents. The battle game in MAgent is a mixed cooperative competition scene. In a pre-defined grid world, two groups of agents are fighting against each other, whose actions are provided by the same or different algorithms. The goal of both groups is to defeat the other, and the group with more remaining agents wins. In each experiment, the two sides play against each other for 50 rounds, and the final outcome is determined based on the win rate. The results of one round will be randomly selected for visualization. Through multiple sets of different experiments, we deeply explore the relationship between emotion and win rate, and prove the effectiveness of the proposed method. Then, the feasibility of the method is further verified by comparing the simulation results with the real scene. And we visualize the results in Unity3D.

\subsection{Comparison of the win rate of both sides}

\subsubsection{Two sides with different emotional states}

The experiment in this section mainly analyzes the influence of the emotion of agents on the battle results for the two sides. Both sides use the same proposed algorithm to provide actions, and the initial number of agent is 256. We set up three sets of experiments to verify the fighting results of the agents under different emotions. The initial emotions of the opposite side are relatively peaceful, and that of the righteous side are respectively peaceful, positive, and negative. Under different emotional states, we randomly assign initial emotional values to each agent. The results of this part of the experiment demonstrated the validity of positive emotions.

\begin{figure}[htbp]
\begin{minipage}{0.5\linewidth}
  \centerline{\includegraphics[height=3.5cm,width=4.5cm]{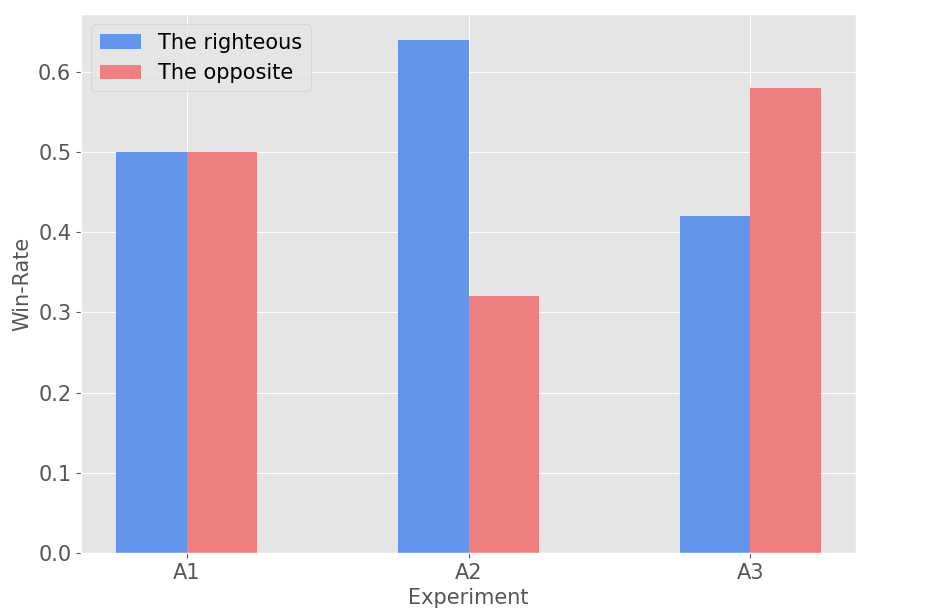}}
  \centerline{(a)}
\end{minipage}
\hfill
\begin{minipage}{.5\linewidth}
  \centerline{\includegraphics[height=3.5cm,width=4.5cm]{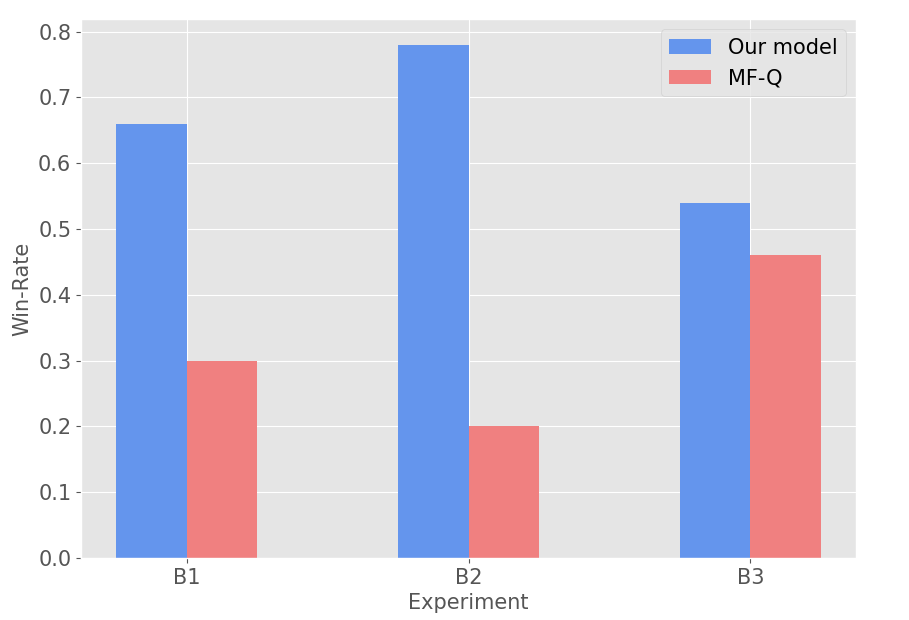}}
  \centerline{(b)}
\end{minipage}

\caption{The comparison of experimental results. (a) The win rate of each side under different emotional states. The blue color represents the righteous and the red represents the opposite. (b) Combat results of two groups using our algorithm and the MF-Q algorithm respectively. The blue color represents our algorithm and the red represents the MF-Q algorithm. }
\label{pic4-1}
\end{figure}

\begin{table}[htbp]
\centering  
\caption{The experimental parameters of two sides with different emotional states.}  
\label{table3}
\begin{tabular}{ccc}
\toprule[1pt]
\begin{tabular}[c]{@{}l@{}} 
Experiment number
\end{tabular} &
\begin{tabular}[c]{@{}l@{}}Number of agent
\end{tabular} & Initial emotional state  \\ \midrule
A1 & 256 vs 256    & \begin{tabular}[c]{@{}c@{}}[0.4,0.6] : [-0.6,-0.4]\end{tabular}      \\
A2 & 256 vs 256    & \begin{tabular}[c]{@{}c@{}}[0.6,1] : [-0.6,-0.4]\end{tabular}      \\
A3 &256 vs 256    & \begin{tabular}[c]{@{}c@{}}[0,0.4] : [-0.6,-0.4]\end{tabular}      \\ \bottomrule[1pt]
\end{tabular}
\end{table}


Figure \ref{pic4-1}(a) shows the battle results of two sides with different emotional states. The experimental parameters are shown in Table \ref{table3}. The second column in this table is the number of agent on the righteous side and that on the opposite side. The third column is the initial emotional state. $T$ in Table \ref{table2} is set to 0.5. According to the results with A1 experimental setting, when other conditions and emotional states are the same, the win rate is basically close. In A2 setting, since the emotional state of the righteous side is more aggressive than that of the opposite side, the win rate is 0.64, winning 32 rounds. In A3 setting, the opposite side achieves a win rate of 0.58. The agents in the righteous group have a negative emotional state and are afraid of the opponent and dare not attack actively. During these three experiments, the sum of the win rates of both groups is not 1.0, it means that there are some draws.

\begin{figure}[htbp]
\centering
\includegraphics[height=4.5cm,width=8.5cm]{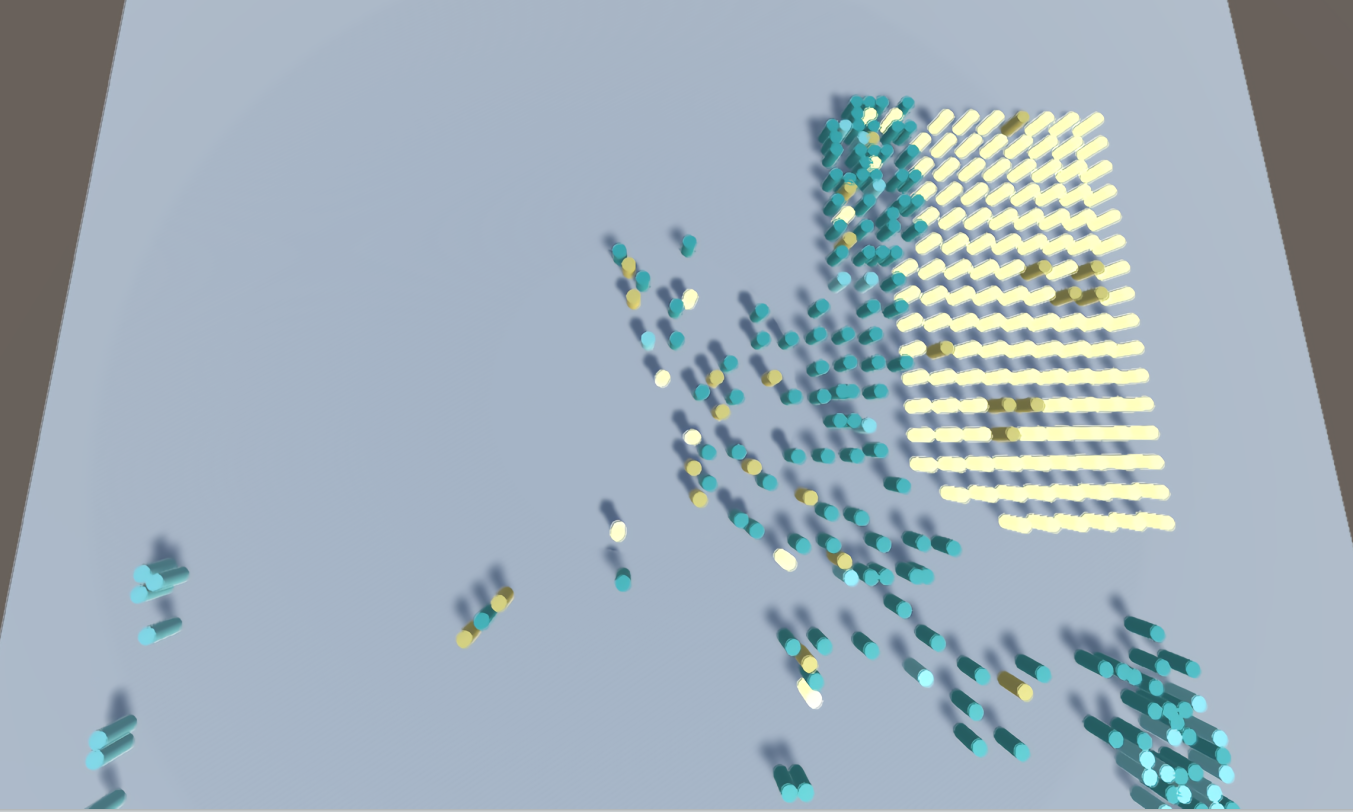}
\centering
\caption{The position of all agents at frame 114. We use the cylinder to represent the agent, where the blue represents the righteous, and the yellow represents the opposite. The darker the color of the cylinder, the more aggressive the combat state.}
\label{pic4}
\end{figure}

From the A2 setting, we randomly select to visualize a round battle, as shown in Figure \ref{pic4}. The darker the color of the righteous indicates the more positive emotions, and the darker the color of the opposite indicates the more negative emotions. It can be seen that a small number of agents on the righteous side are conservative in light blue, and some agents on the opposite side are aggressive in dark yellow. Agents with darker colors are in aggressive state and tend to attack the opponents, most of which are located in areas where the two sides fight fiercely. Agents with lighter colors are in conservative combat state. They are afraid of the opponent and are unwilling to attack the opponent, and most of them are floating on the edge of the scene.

The simulation in this section proves that when the righteous is fighting the mob, it should have positive emotions and maintain the enthusiasm for combat. In an aggressive combat state, it will be more inclined to take offensive actions, have a greater probability of subduing the opponent, and improve the combat win rate to a certain extent.

\subsubsection{Two sides with different algorithms}
In this section, we compare our algorithm with the MF-Q \cite{16} algorithm and our previous ACSEE \cite{li2019acsee} algorithm. We compare the win rate of the ACSED with the original MF-Q algorithm, and then, the ratios of the remaining numbers of the two sides of the three algorithms under the same conditions are compared. When other experimental conditions are the same, we try to figure out which algorithm the group adopts will have a higher battle win rate. At the same time, the rationality and reality of the simulation results of the three algorithms are judged.

To compare win rates with the original MF-Q algorithm, we set up three sets of experiments. The righteous side uses the algorithm proposed in this paper, the agent can perceive the emotions of itself and others, and they can plan their actions based on the information they perceive. The opposite side uses the MF-Q algorithm, planning action without considering individual emotional factors. The agent does not have emotional attributes and will not be able to perceive relevant information.



The results of the three sets of experiments are shown in Figure \ref{pic4-1} (b). In the B1 experiments, the emotional state of both groups is the same. Three comparative experiments are conducted for different emotional states of peaceful, positive and negative. The mean value of the three experimental results are calculated. The results show that the proposed algorithm in this paper is better than the MF-Q algorithm. The win rate is higher. In the B2 experiments, the emotional state of both groups is positive, that means the righteous is more aggressive. The results of the match are shown in the second row of the table. The righteous have a win rate of 78\%, winning 39 rounds, and surpassing the opponents. In the B3 experiments, the emotional state of both groups is negative. The righteous are afraid of the opposite, and the opponents tend to provoke and attack the righteous. In this case, the opposite can take an advantage of the negative emotions and the opponent's low morale to attack the righteous side. This will have a higher win rate. However, as shown in the third row of the table, the win rate of both groups is 0.54 : 0.46, the winning number of righteous is slightly more. Since the opposite agent cannot perceive the emotions of itself and others, it cannot use the emotional advantage to plan favorable actions. When the righteous is planning action, they can be combined with emotional information for comprehensive consideration. If the righteous perceive that the opponent's emotional state is negative and the combat state is aggressive, they will tend to adopt moving-type actions to avoid blind attacks and protect themselves from harm. Therefore, the opposite side has a low win rate.

\begin{figure}[!t]

\begin{minipage}{0.48\linewidth}
  \centerline{\includegraphics[height=3.4cm,width=4.4cm]{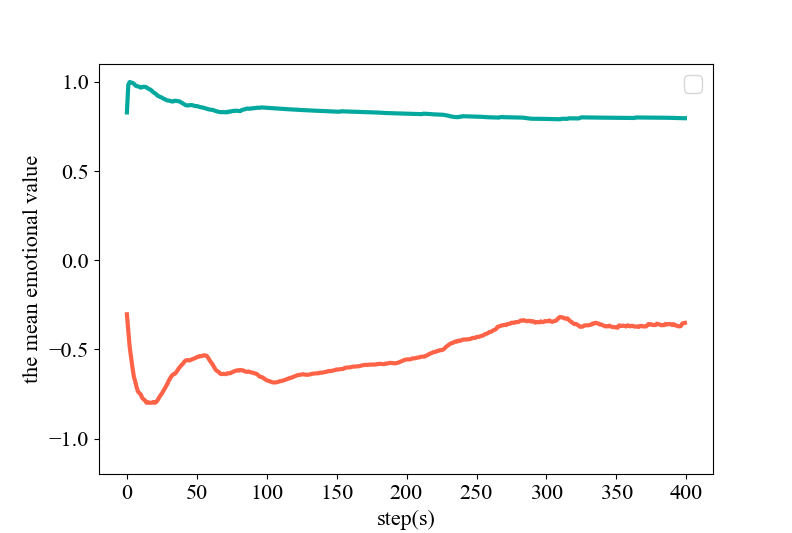}}
  \centerline{(a)}
\end{minipage}
\hfill
\begin{minipage}{.48\linewidth}
  \centerline{\includegraphics[height=3.4cm,width=4.4cm]{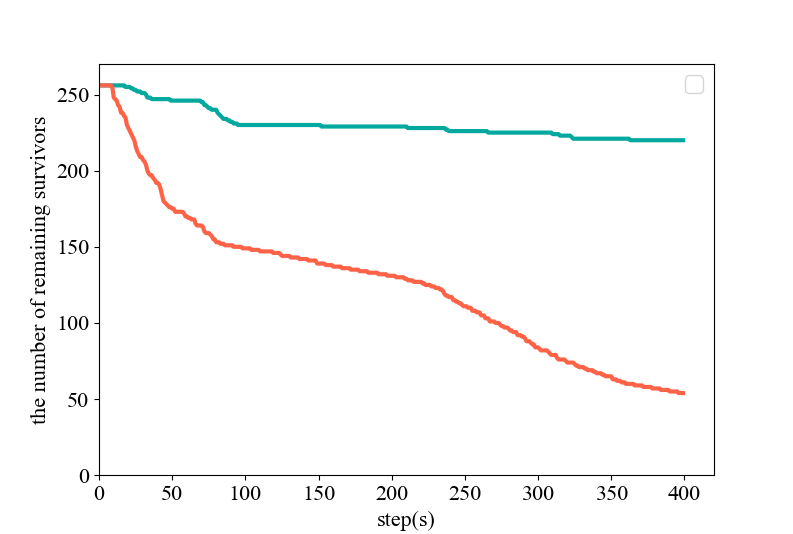}}
  \centerline{(b)}
\end{minipage}

\caption{The changes in emotional value and the number of remaining survivors of both groups at each time step, when the initial emotional state of the righteous and the opposite is positive. This means that the righteous tend to be aggressive, while the opposite tend to be conservative and defensive. The blue curve represents the righteous and the red curve represents the opposite. (a) represents the change in the mean emotional value of the members of both groups; (b) represents the remaining number of surviving members of both groups.}                              
\label{pic5}
\end{figure}

\begin{figure}[!t]

\begin{minipage}{0.48\linewidth}
  \centerline{\includegraphics[height=3.4cm,width=4.4cm]{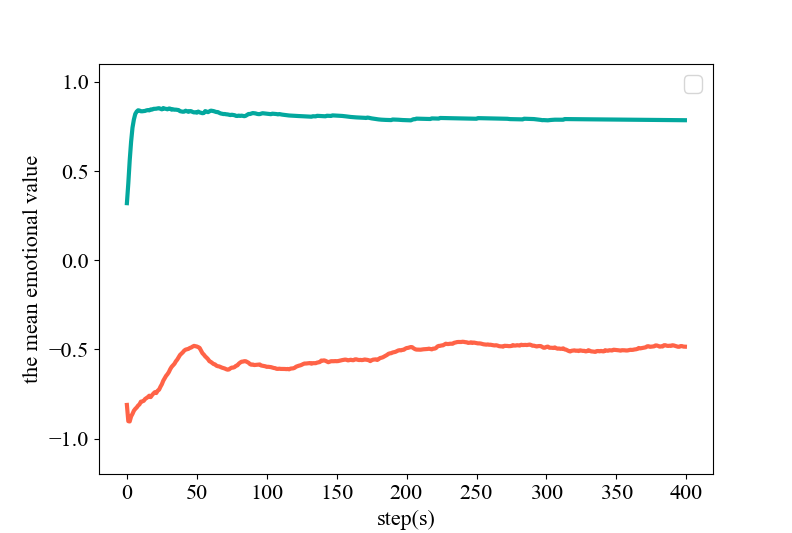}}
  \centerline{(a)}
\end{minipage}
\hfill
\begin{minipage}{.48\linewidth}
  \centerline{\includegraphics[height=3.4cm,width=4.4cm]{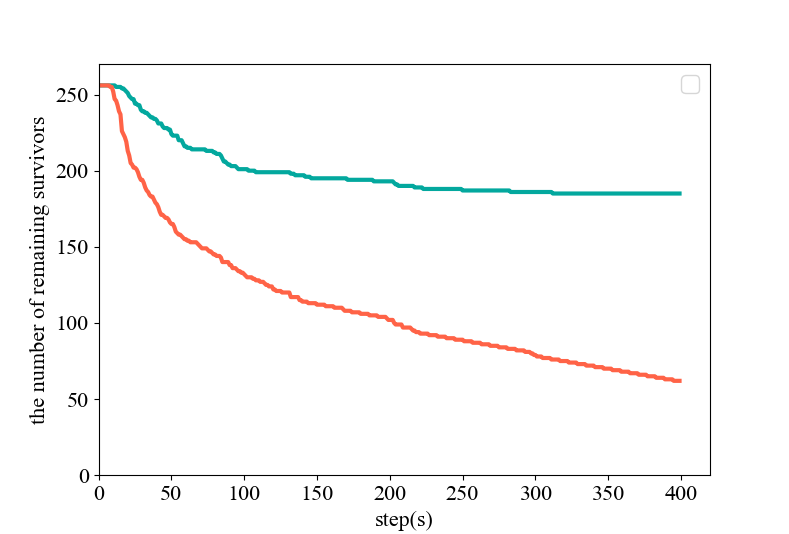}}
  \centerline{(b)}
\end{minipage}

\caption{The changes in emotional value and the number of remaining survivors of both groups at each time step, when the initial emotional state of the righteous and the opposite is negative. This means that the righteous tend to be conservative and defensive, while the opposite tend to be aggressive. The blue curve represents the righteous, and the red curve represents the opposite side. (a) represents the change in the mean emotional value of the members of both groups; (b) represents the remaining number of surviving members of both groups.}
\label{pic6}
\end{figure}

We randomly select a round of battles from the B2 and B3 experiments, as shown in Figures \ref{pic5} and \ref{pic6}. The blue curve represents the righteous and the red curve represents the opposite. Figure \ref{pic5}(a) is the mean emotional value of both groups at each step. The initial emotional state of the righteous and the opposite are both positive. As the battle progresses, the emotions of the righteous become more positive, and then remain in a positive state. The emotions of the opposite firstly gradually become more negative and then gradually become positive. Figure \ref{pic5}(b) is the remaining number of survivors on both sides at each step. At about step 200, as the emotions of the opposite gradually become more positive, the agent becomes more afraid of the righteous and dare not fight, resulting in a large number of members are subdued by the righteous, and the remaining number of survivors is rapidly decreasing. In Figure \ref{pic6}(a), the initial emotional state of the righteous is negative. However, as the battle progresses, the emotions of the righteous quickly become positive, and the action is adjusted in time based on the perceived emotional information, and eventually win. Figure \ref{pic6}(b) shows that at the end of the battle, the number of survivors on the righteous side is significantly greater than that on the opposite side.

We use a new indicator to compare the three algorithms of ACSED, MF-Q, and ACSEE. This indicator is the ratio of the number of the righteous and the opposite, indicating the ratio of the number of survivors on both sides at the same time. In the experiments, we set the same experimental conditions, including the location of the agent, the number of people, the emotion, the size of the map, and so on. We ensure the consistency of the experimental conditions, and then use the three algorithms for the adversarial crowd simulation and record the number of survivors in each algorithm at each time step. When setting the initial emotional value, considering that the main task of the righteous side in the riot scene is to pacify the riot, the righteous emotion is designed to be a more positive emotion, which tends to actively attack the other group. The overall quality of the righteous is higher than the opposite, and the will to subdue the enemy is stronger. Thus, we set the emotional value range of the righteous is (0.7, 0.9), the emotional value range of the opposite is (-0.8, 0.5). In the initial state, there are equal numbers on both sides.

\begin{figure}[!t]
\centering
\includegraphics[height=6.5cm,width=8.5cm]{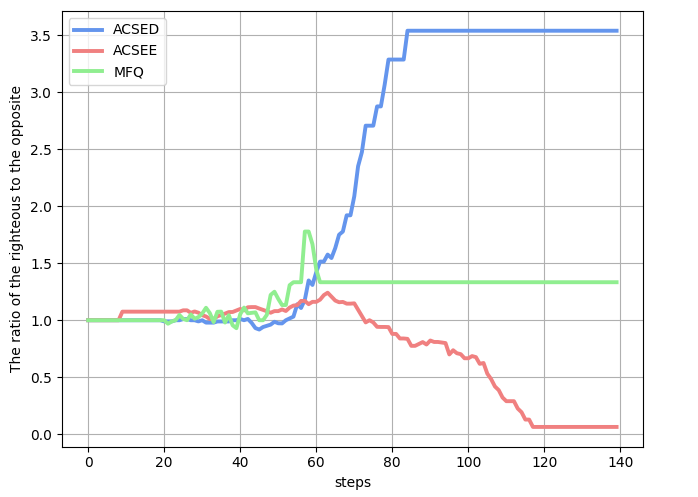}
\centering
\caption{The changes in the ratio of the number of the righteous and the opposite under different algorithms, and the ratio varies with time step.}
\label{pic9}
\end{figure}

We randomly select a round of battles, as shown in Figure \ref{pic9}. According to the experimental results, the MF-Q algorithm has the fastest convergence speed, which is slightly faster than our algorithm. However, the MF-Q algorithm does not consider the emotion of the agent, and cannot perceive the emotional changes of the surrounding teammates and opponents. This will simplify the battle process to a certain degree, reduce the rationality of the simulation, thereby speeding up the convergence process. In our results, the righteous overwhelm more opponents with less cost. Under the same conditions, the righteous fails in the ACSEE algorithm. Although the ACSEE algorithm considers the antagonistic emotions of the crowd, it does not consider the situation where the two sides face each other first and then battle. In the riot, not only the two sides battle directly, but also the two sides face each other first and then battle. The simulation scope of the ACSEE algorithm is more local. So in a round of random experiments, the righteous fails. Under the indicator of the ratio of the number of survivors of righteous and opposite sides, the results of our algorithm are significantly higher than the other two algorithms. While ensuring that the crowd's emotions are taken into account, in our algorithm, the righteous side can subdue the other side faster, and the ratio is increased from the initial 1 to 60, which shows that more opponents are subdued. More details can be seen in the supplementary video.

Experiments have proved that by fully considering the emotions of both sides we can truly grasp the current battle situation, understand the opponent's combat state and the actions taken. It proves the rationality and necessity of considering emotional factors when planning the action of both sides. In the event of a sudden riot, using the algorithm proposed in this paper provides help for the righteous to formulate better uniform strategies.

\subsubsection{Two sides with different numbers of agents}
\begin{figure}[htbp]
\centering
\includegraphics[height=5cm,width=8cm]{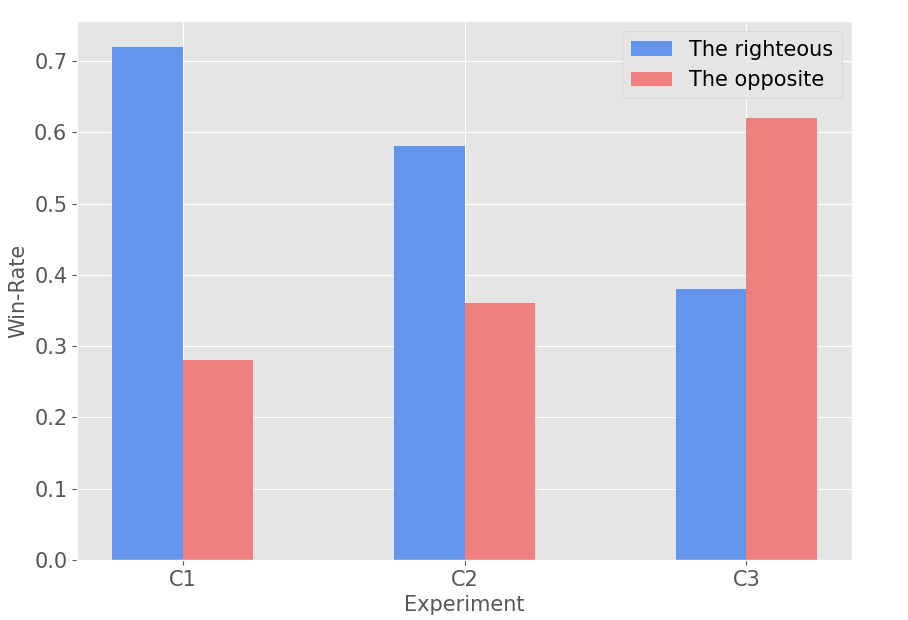}
\centering
\caption{The win rate of both sides under different numbers of people using our ACSED algorithm. In each experiment, the blue represents the righteous and the red represents the opposite.}
\label{pic4.3}
\end{figure}

The experiment in this section discusses the influence of different numbers of people on combat results when the righteous and the opposite emotional states are both positive. According to Table \ref{table2}, the positive emotion on both sides means that the righteous side is more aggressive than the opposite side. This section consists of two sets of experiments. In the first set of experiments, there are three comparative experiments, the number of combatants on the righteous side remains unchanged, and the number of the opposite side is changed. Both sides use the proposed algorithm in this paper to fight. The initial number is: 75 vs 75, 75 vs 90, 75 vs 100. Figure \ref{pic4.3} is the result of the experiments. When the number of both sides is same, the righteous wins. When the initial number of the opposite is increased to 90, the righteous still win. When the number is increased to 100, the righteous lose. Experiments show that the righteous side can win more with less when the two sides have positive emotions, but when the initial number of the two sides is quite different, the probability of winning will become less.

\begin{table}[htbp]
\centering  
\caption{The win rate of both sides under different numbers of people using the ACSED algorithm.}  
\label{table5}
\begin{tabular}{cc}
\toprule[1pt]
Initial number of both groups & Win rate                      \\ \midrule
75 : 75                        & 0.72 : 0.28                     \\
75 : 90                        & 0.58 : 0.36                     \\
75 : 100                        & 0.38 : 0.62 \\ \bottomrule[1pt]
\end{tabular}
\end{table}

\begin{table}[htbp]
\centering  
\caption{The winning rate of both sides with different numbers of people using different algorithms.}  
\label{table6}
\begin{tabular}{cc}
\toprule[1pt]
Initial number of both groups & Win rate                      \\ \midrule
256 : 256                        & 0.78 : 0.2                    \\
192 : 256                        & 0.62 : 0.38                     \\
128 : 256                        & 0.34 : 0.62 \\ \bottomrule[1pt]
\end{tabular}
\end{table}

Table \ref{table6} is the result of the second set of experiments. The righteous use ACSED algorithm and the opposite use MF-Q algorithm. When the initial number of the righteous is 256 and 192, the righteous win. When the initial number of the righteous is reduced to 128, the righteous lose the battle. The algorithm proposed in this paper can still win more with less when playing against the MF-Q algorithm, and the righteous still can't beat the opposite when the initial number of people gap is large. Moreover, in these two experiments where the righteous win in Table \ref{table6}, the win rate is generally higher than that of the righteous in Table \ref{table5}.

Experiments prove that the righteous side who has a positive emotional state in combat can increase the win rate to a certain extent. Subject to objective conditions, if there is a large difference in the number of people between the two sides in the battle, it cannot rely solely on the positive emotional state to win. In the face of real crowd confrontation incidents, the experimental conclusions in this section have some reference value for formulating actual combat plans. When formulating plans, it is necessary to combine emotions and reasonably arrange the number of participants in combat.

\begin{figure*}[t]
\centering
\subfigure[A real crowd antagonistic scene]{
\centering
\includegraphics[height=3.5cm,width=17.2cm]{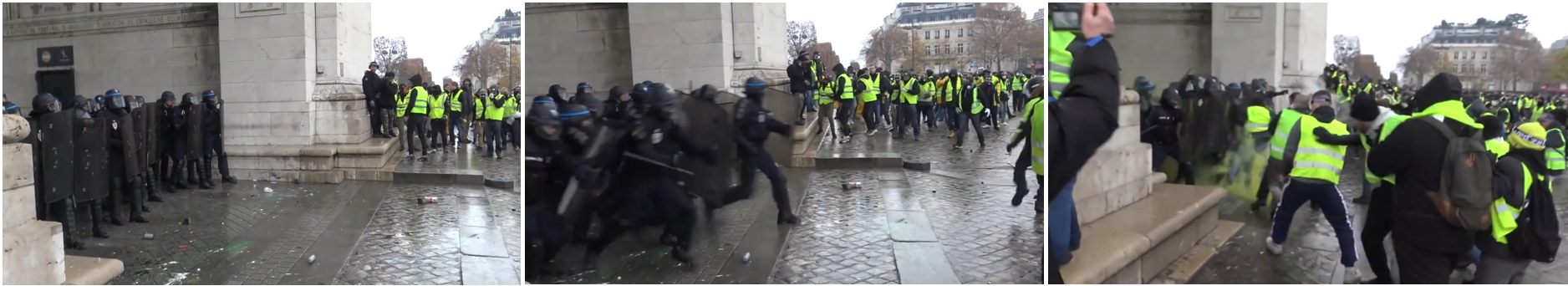}
}%
\quad
\subfigure[Our simulation results]{
\centering
\includegraphics[height=3.5cm,width=17.2cm]{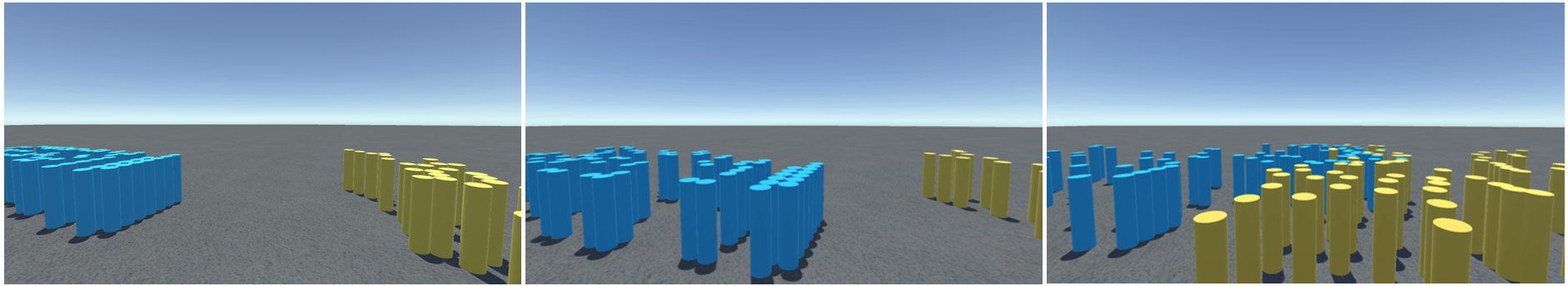}
}%
\quad
\subfigure[A real confrontation exercise scene]{
\centering
\includegraphics[height=3.5cm,width=17.2cm]{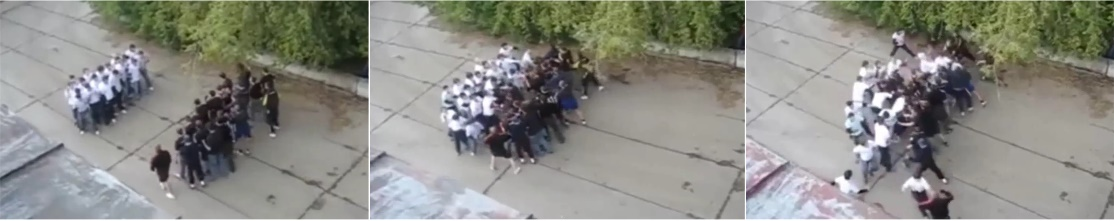}
}%
\quad
\subfigure[Our simulation results]{
\centering
\includegraphics[height=3.5cm,width=17.2cm]{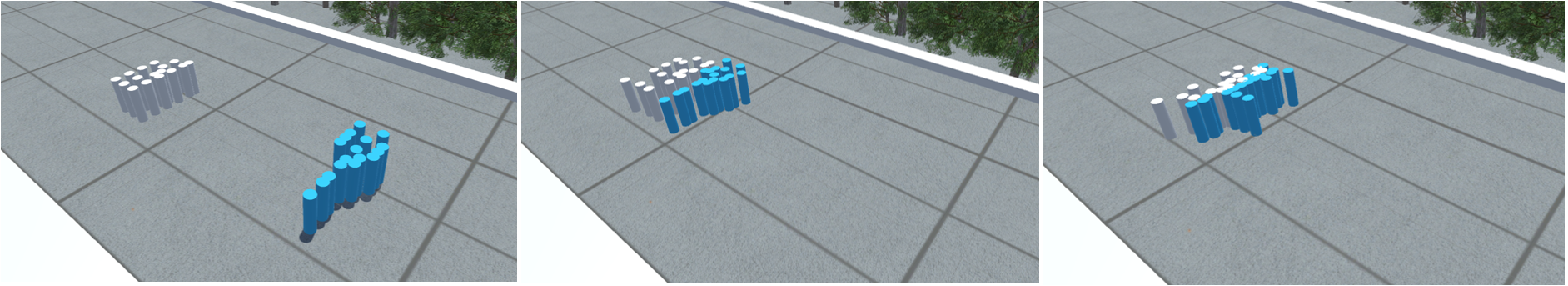}
}%
\centering
\caption{Comparisons between real scenes and our simulation results. (a) is a real crowd antagonistic scene. (b) is our corresponding simulation results. The yellow cylinder represents the group wearing black who is the opposite crowd, and the blue cylinder represents the group wearing black clothes who is police. (c) is a real confrontation exercise scene. (d) is our corresponding simulation results. The white cylinder represents the group wearing white clothes, and the blue cylinder represents the group wearing dark clothes.}
\label{pic7}
\end{figure*}

\subsection{Comparison of simulation results with real scenes}

In this section, we use the proposed algorithm to simulate real confrontation cases, and verify the authenticity and practicability of the simulation results through crowd movement trends and position distribution. Besides, we compare our algorithm simulation result with the MF-Q algorithm and the ACSEE algorithm. The main goal of our algorithm is to predict the movement trend of the crowd, and under certain mission constraints, formulate better combat strategies. The results show that our algorithm can obtain simulation results that are the closest to real video.




Figure \ref{pic7}(a) is a real riot. The purpose of the police is to control the riots provoked by the lawless, and to subdue them to maintain social order. In the video of the real riot scene, the police have two main tasks, including subduing the lawless and preventing them from passing through the gate in the image. First of all, under the condition of constant provocation by the lawless, the police take the initiative to attack the lawless. The lawless constantly move and retreat because of the fear of the police. Then, after the police rush out a distance, they retreat to defend the gate in order to carry out the second task. Finally, the lawless see the police retreat, and immediately rush to the police to attack, so that the two sides fight. In order to make our simulation results more realistic and reasonable, only local control is not enough, so global control is also added. Therefore, the task of the police is abstracted into global control, with rewards for subduing the lawless and defending gates. It is necessary to change the parameter setting of the reward. Suppose the gate is a region in the grid world. If the distance between the righteous member and this area is greater than a certain threshold, it will be punished at the current time step. Obviously, the righteous retreat after the attack in the real video. Therefore, we add the rule that when the opposite retreat in a large scale, the righteous also retreat to defend the gate. The final simulation results are shown in Figure \ref{pic7}(b). More details can be seen in the supplementary video.

Figure \ref{pic7}(c) is a real confrontation exercise scene. The group wearing white clothes represents the righteous, and the group wearing black clothes represents the opposite. At first the opposite attack the righteous, and then they begin to battle. The two sides are evenly matched in numbers and the behavior of the opposite is still under control, so the righteous choose not to take the initiative to attack. This requires the addition of global control. The righteous must obey the overall combat strategy and cannot rush to the opponent just because of aggressive emotional state. Therefore, we add global control as a rule to limit the actions of the righteous. When the distance between the two sides is lower than a certain threshold and the emotional state of the opposite is very aggressive, the righteous will start to attack. Figure \ref{pic7}(d) is a simulated scene. Details are in the supplementary video.

We quantitatively evaluate the simulation results using dominant path, entropy metric and angular error as \cite{li2019acsee}. The dominant path is defined based on the collectiveness of the movement of the crowd. Collectiveness describes the extent to which individuals act as a unit in collective movements and is a fundamental and pervasive measure of various crowd systems, including crowds in confrontation scenarios. We calculate collectiveness using the method in \cite{7990566}. A group is formed when the collectiveness of the agents in a certain area is significantly higher than that of the surrounding area. The center of this group is determined from the average of the positions of all agents in the group. The trajectory at the center of the group forms the dominant path. We use this method to calculate the dominant path of the real-world video and our simulation results. We then evaluate our crowd simulation results using entropy metric and angular error. They are used to evaluate the error of the trajectory and movement direction, respectively.

The entropy metric is employed to evaluate the error between the dominant path of the simulation results and that of the real-world video. The lower the entropy value, the higher the similarity. It can be seen from the Table \ref{table7} that our algorithm has the lowest entropy value and is closest to the dominant path of the real video. We also use the angular error between the movement direction in the simulation results and that in the real video as an evaluation metric. Angular error is defined in Formula \eqref{equ16}. $V_{x}$ and $V_{y}$ is the movement directions of the simulation results. $V_{x g t}$ and $V_{y g t}$ is the movement direction of the real-world video.

\begin{scriptsize}
\begin{equation}
\label{equ16}
AE = \cos ^{-1}\left(\left(V_{x} \cdot V_{x g t}+V_{y} \cdot V_{y g t}\right) / \sqrt{V_{x}^{2}+V_{y}^{2}} \sqrt{V_{x g t}^{2}+V_{y g t}^{2}}\right)
\end{equation}
\end{scriptsize}


\begin{table}[h]
\centering
\caption{Entropy metric and angular error of three simulation algorithms under different scenarios.}  
\label{table7}
\scriptsize
\renewcommand{\arraystretch}{1.4}
\setlength{\tabcolsep}{3mm}{
\begin{tabular}{|cl|c|c|c|}
\hline
\multicolumn{2}{|c|}{Algorithm}                                                                          & ACSED & MF-Q & ACSEE \\ \hline
\multicolumn{1}{|c|}{\multirow{2}{*}{\begin{tabular}[c]{@{}c@{}}Entropy\\ metric\end{tabular}}} & scene1 &  0.7877     &  0.952    &   1.151    \\ \cline{2-5} 
\multicolumn{1}{|c|}{}                                                                          & scene2 &  0.1330     &  0.178    &   0.189    \\ \hline
\multicolumn{1}{|c|}{\multirow{2}{*}{\begin{tabular}[c]{@{}c@{}}Angular\\ error\end{tabular}}}  & scene1 &  0.242/0.311    &  0.512/0.673    &  0.487/0.543     \\ \cline{2-5} 
\multicolumn{1}{|c|}{}                                                                          & scene2 &  0.189/0.013     &  0.219/0.612    &  0.398/0.016     \\ \hline
\end{tabular}}

\end{table}

A lower entropy value means a higher similarity between simulation results and real-world scenarios. Simulations with entropy value less than 1.0 are considered visually very similar to the source data, while those with value greater than 6.0 are visually very different. A lower value of angular error means a higher similarity to real-world crowd videos. We report the mean and variance of angular error at different time steps. Table \ref{table7} shows that our algorithm consistently outperforms the MF-Q algorithm and the ACSEE algorithm. Compared with the MF-Q algorithm, our algorithm takes into account the influence of emotions on the agent and can formulate a combat strategy with a higher win rate. Compared with the ACSEE algorithm considering the evolutionary game theory, we use deep reinforcement learning to predict adversarial behavior, and the simulation results obtained are more reasonable.

By comparing the real scene and the simulation one, we can see that the movement trend of the crowd in the simulation is basically the same as that in the real scene. This shows that our algorithm is in line with reality and can simulate crowd confrontation incidents well. 

\section{Conclusion}
This paper proposes an emotional contagion-aware deep reinforcement learning model for antagonistic crowd simulation. We build an antagonistic emotional contagion module based on the SIS epidemic model and the reward value obtained by the agent according to the combat situation. When modeling crowd behaviors, the deep reinforcement learning technology is used to predict the action of the agent more closely to human thinking. The DQN and mean field theory are introduced to predict the action in the crowd. In addition, our model considers the specific influence of emotions on antagonistic crowd behaviors. We determine the individual combat state through the emotional value, and re-plan more reasonable actions for the agent according to the behavior rules. The model proposed in this paper is proved through a variety of experiments. For one thing, it can simulate the antagonistic crowd behaviors more realistically and help to study crowd movement trend under riots. For another, it can provide a reference for the formulation of combat plans, thereby improve the win rate.

Although our work can contribute to the control and calming of emergencies, there are still some shortcomings. In real world, groups that provoke riots are often very irrational and extreme, they are more uncontrollable than normal groups. The simulation result of our model may only be one case of many real situations, and they may not completely follow the result calculated by our model. In future work, we will continue to improve our prediction results considering more actual situations of antagonistic crowds. Furthermore, we cannot directly obtain or accurately infer the emotions in crowd antagonistic scenes. The videos are generally captured by passers-by at the riot scene, and they will shake the video and cause the picture to be chaotic because of their fear and nervousness. The quality of most of the real videos is usually poor. Therefore, the initial emotional state of our model is empirically set from real-world videos, which is not very accurate. In the future, we plan to use the latest wearable devices to collect this data, providing an efficient way to obtain the initial state of our model more efficiently and accurately. 

\ifCLASSOPTIONcaptionsoff
  \newpage
\fi

\begin{IEEEbiography}[{\includegraphics[width=1in,height=1.25in,clip,keepaspectratio]{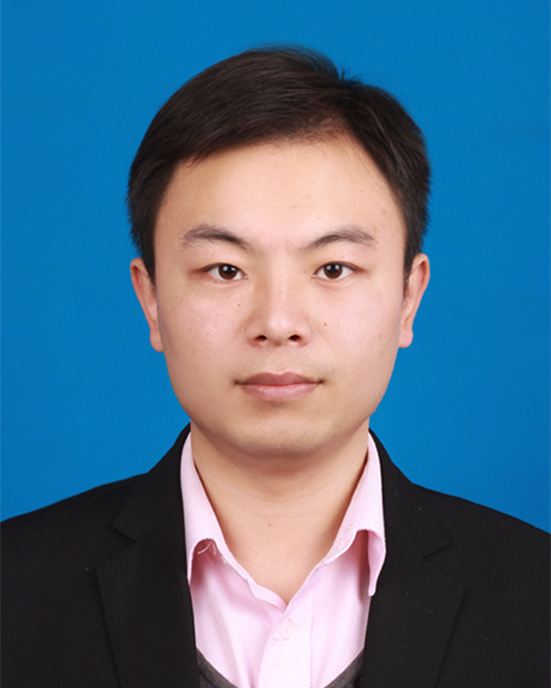}}]
{Pei Lv} received the Ph.D. degree from the State Key Laboratory of CAD\&CG, Zhejiang University Hangzhou, China, in 2013. He is an Associate Professor with the School of Computer and Artificial Intelligence, Zhengzhou University, Zhengzhou, China. His research interests include computer vision and computer graphics. He has authored more than 30 journal and conference papers in the above areas, including the IEEE T RANSACTIONS ON IMAGE PROCESSING , the IEEE TRANSACTIONS ON CIRCUITS AND SYSTEMS FOR VIDEO TECHNOLOGY, the IEEE TRANSACTIONS ON AFFECTIVE COMPUTING, CVPR, ACM MM, and IJCAI.
\end{IEEEbiography}



\begin{IEEEbiography}[{\includegraphics[width=1in,height=1.25in,clip,keepaspectratio]{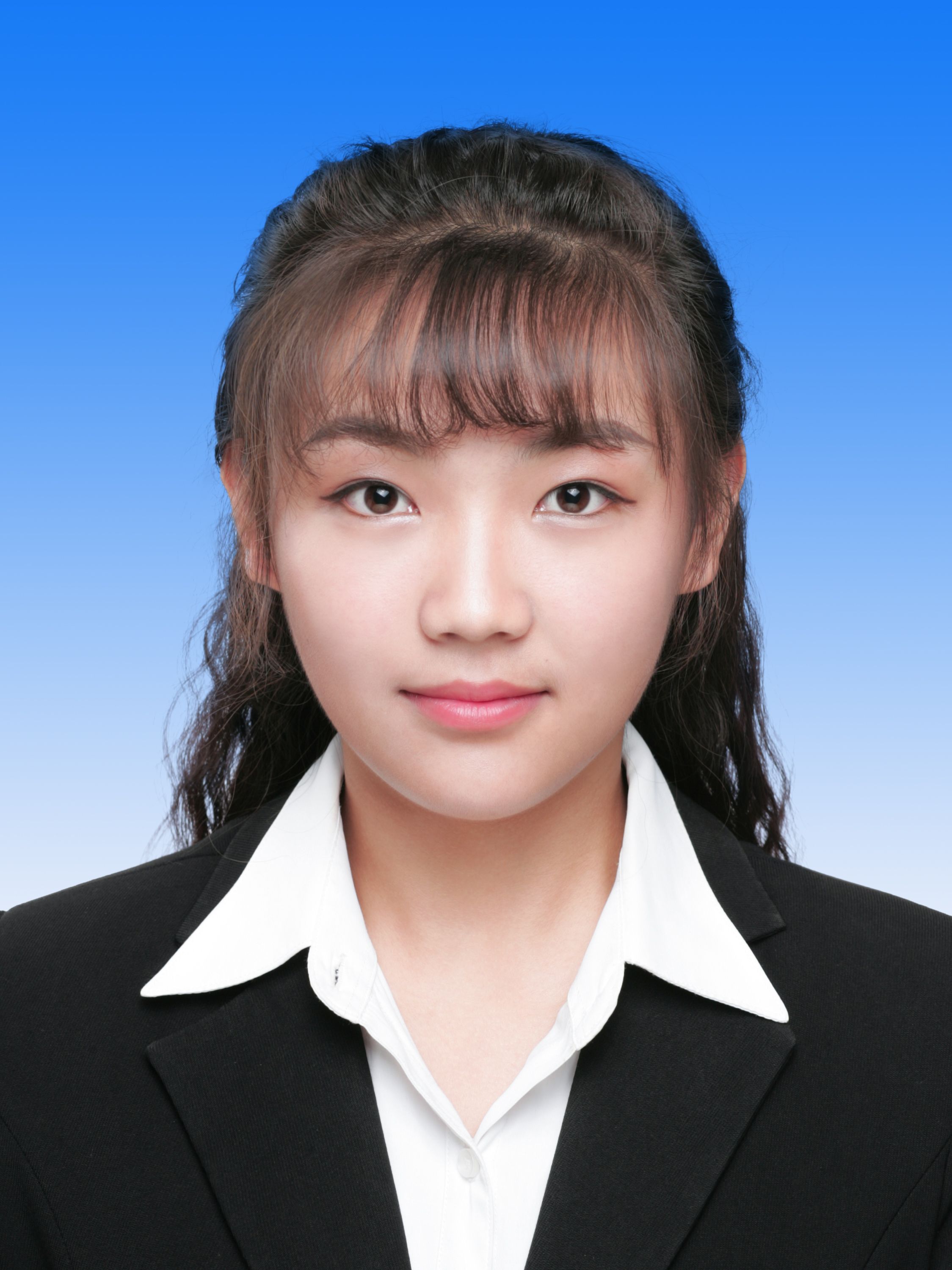}}]
{Qingqing Yu} received the B.S. degree from the Digital Media Technology Department, Shandong University, China, in 2020. She is currently a master student in the School of Computer and Artificial Intelligence of the Zhengzhou University. Her research interests include computer graphics, crowd simulation.
\end{IEEEbiography}

\begin{IEEEbiography}[{\includegraphics[width=1in,height=1.25in,clip,keepaspectratio]{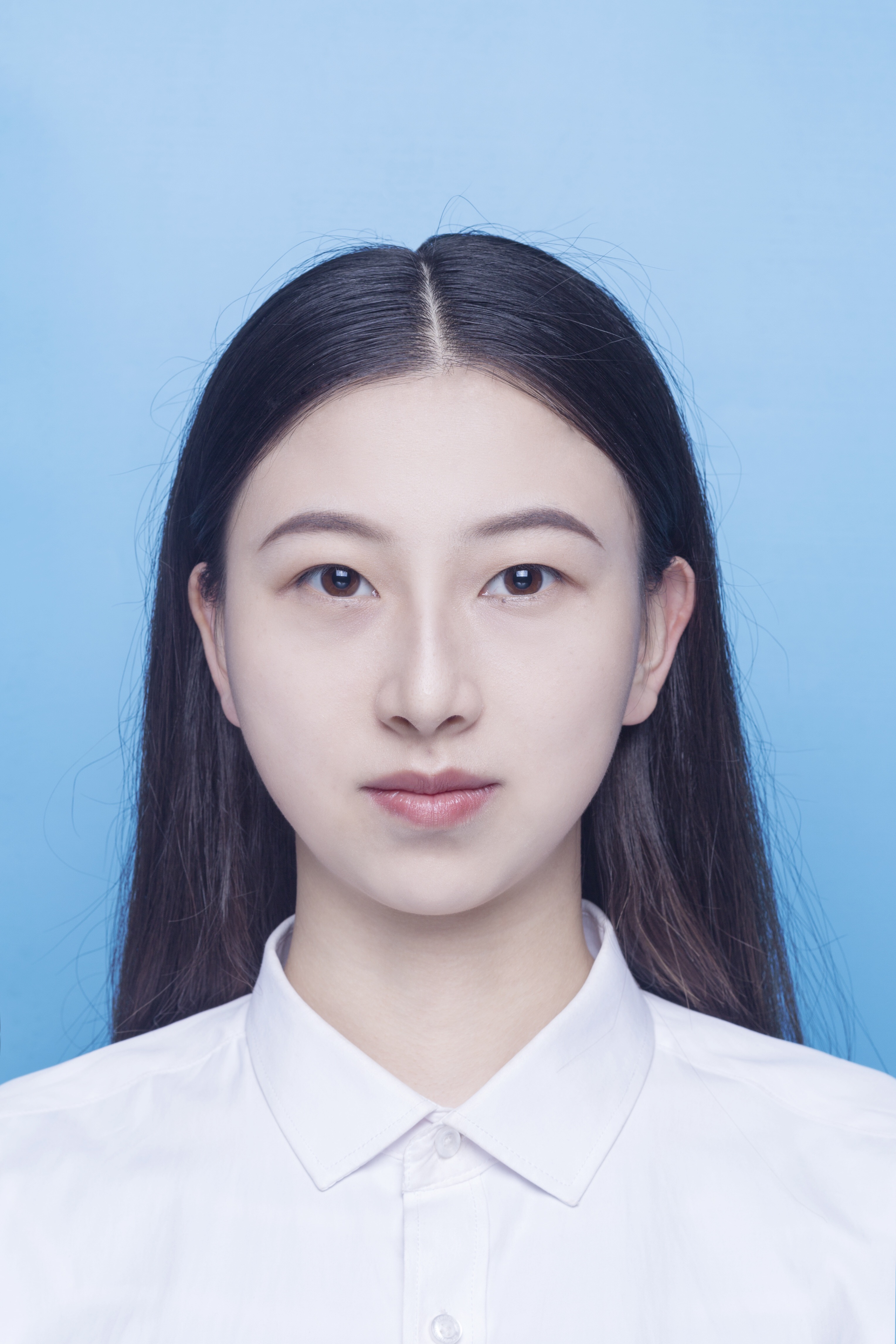}}]
{Boya Xu} received the M.S. degree from the School of Information Engineering (2021), Zhengzhou University, China. Her research interests include computer graphics, crowd simulation.
\end{IEEEbiography}


\begin{IEEEbiography}[{\includegraphics[width=1in,height=1.25in,clip,keepaspectratio]{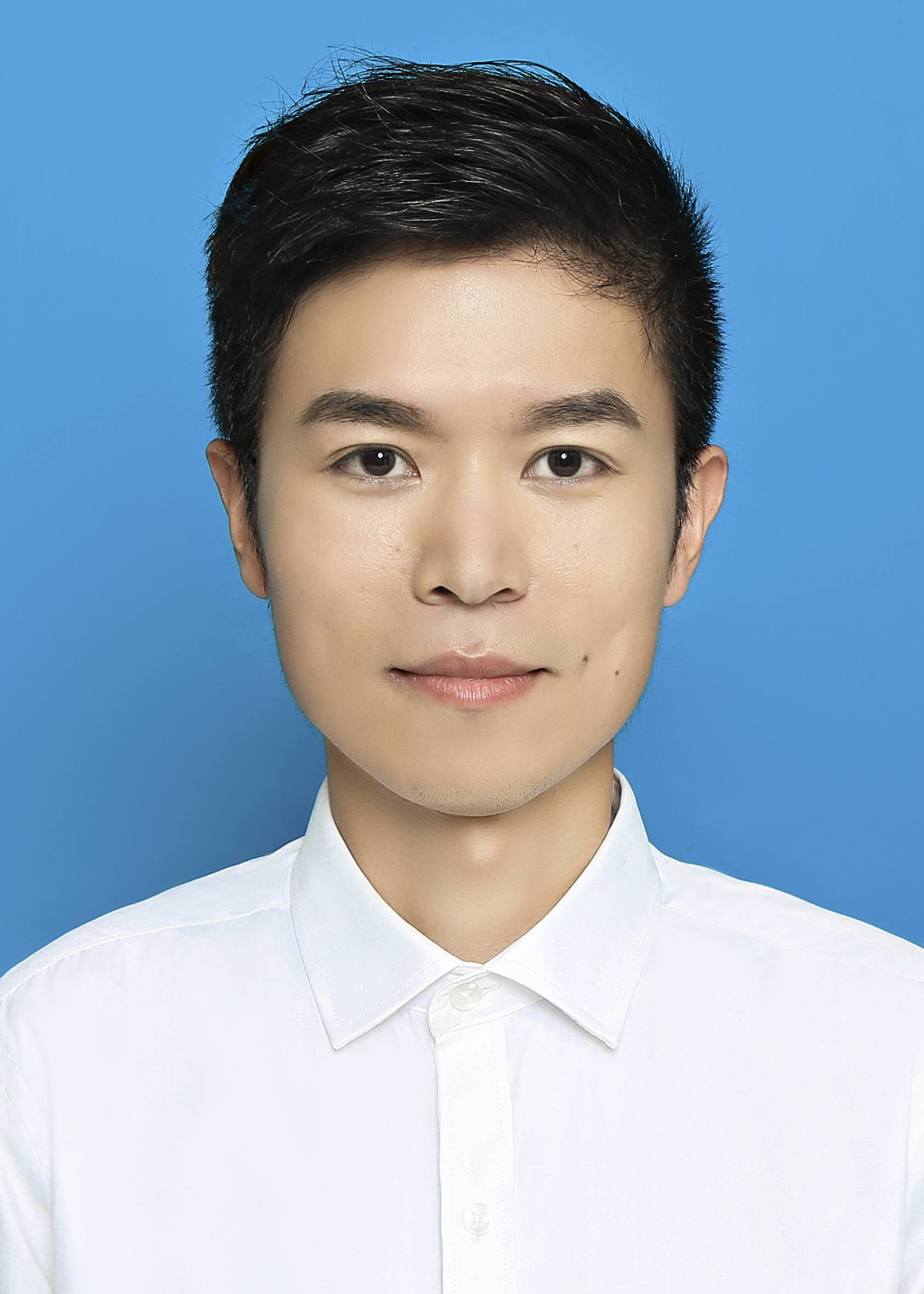}}]
{Chaochao Li} received his Ph.D. degree from the School of Information Engineering, Zhengzhou University, Zhengzhou, China. His current research interests include computer graphics and computer vision. He is currently an assistant research fellow with the School of Computer and Artificial Intelligence, Zhengzhou University, Zhengzhou, China. He has authored over 6 journal and conference papers including the IEEE TRANSACTIONS ON AFFECTIVE COMPUTING, IEEE TRANSACTIONS ON INTELLIGENT TRANSPORTATION SYSTEMS, and IEEE TRANSACTIONS ON SYSTEMS, MAN, AND CYBERNETICS: SYSTEMS.
\end{IEEEbiography}

\begin{IEEEbiography}[{\includegraphics[width=1in,height=1.25in,clip,keepaspectratio]{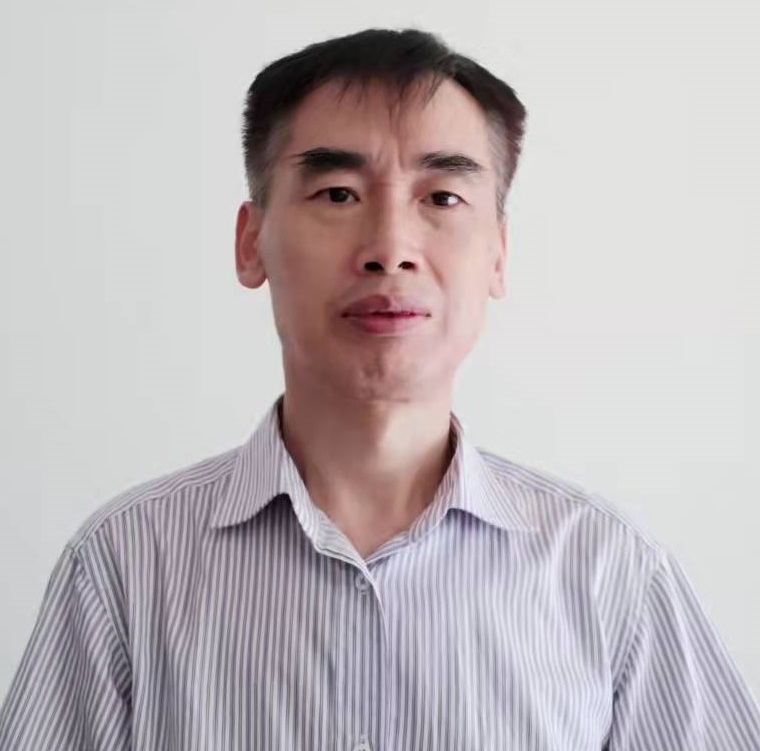}}]
{Bing Zhou} received the B.S. and M.S. degrees in computer science from Xian Jiao Tong University, Xian, China, in 1986 and 1989, respectively, and the Ph.D. degree in computer science from Beihang University, Beijing, China, in 2003. He is currently a Professor with the School of Computer and Artificial Intelligence, Zhengzhou University, Zhengzhou, China. His research interests include video processing and understanding, surveillance, computer vision, and multimedia applications.
\end{IEEEbiography}

\begin{IEEEbiography}[{\includegraphics[width=1in,height=1.25in,clip,keepaspectratio]{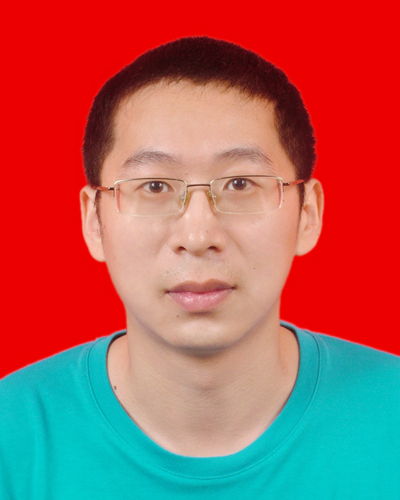}}]
{Mingliang Xu} received the Ph.D. degree in computer science and technology from the State Key Laboratory of CAD\&CG, Zhejiang University, Hangzhou, China, in 2012. He is a Full Professor and the Director with the School of Computer and Artificial Intelligence, Zhengzhou University, Zhengzhou, China. His research interests include computer graphics, multimedia, and artificial intelligence. He has authored more than 100 journal and conference papers in the above areas, including the ACM Transactions on Graphics, the ACM Transactions on Intelligent Systems and Technology, the IEEE T RANSACTIONS ON P ATTERN A NALYSIS AND M ACHINE I NTELLIGENCE , the IEEE T RANSACTIONS ON I MAGE P ROCESSING , the IEEE T RANSACTIONS ON C YBERNETICS , the IEEE T RANSACTIONS ON C IRCUITS AND S YSTEMS FOR V IDEO T ECHNOLOGY , ACM SIGGRAPH (Asia), ACM MM, and ICCV.
	
\end{IEEEbiography}





\bibliographystyle{IEEEtran}
\bibliography{new}

\end{document}